\documentclass[letterpaper]{article} 
\usepackage[preprint]{aaai2027}  
\usepackage[hyphens]{url}  
\usepackage{graphicx} 
\usepackage{natbib}  
\usepackage{caption} 
\usepackage{algorithm}
\usepackage{algorithmic}

\usepackage{newfloat}
\usepackage{listings}
\usepackage[lining]{carlito}
\usepackage[listings]{tcolorbox}
\DeclareCaptionStyle{ruled}{labelfont=normalfont,labelsep=colon,strut=off} 
\floatstyle{ruled}
\newfloat{listing}{tb}{lst}{}
\floatname{listing}{Listing}

\usepackage{booktabs}

\usepackage{amsmath,amssymb}
\usepackage{multirow,colortbl}

\definecolor{controlled}{gray}{0.91}

\title{PI-Mem: Pushing Long-Context Reasoning to 3.6M Tokens\\with Parallel-Iterative Memory}
\author{
    Dawei Liu\textsuperscript{\rm 1,\rm 2}\equalcontrib,
    Haixu Song\textsuperscript{\rm 2,\rm 3}\equalcontrib,
    Shuang Cheng\textsuperscript{\rm 2,\rm 4},
    Shijie Wang\textsuperscript{\rm 2},
    Haozheng Hou\textsuperscript{\rm 2,\rm 5},
    Kaifeng Liu\textsuperscript{\rm 2,\rm 5},\\
    Ermo Hua\textsuperscript{\rm 2,\rm 3},
    Zhonghang Yuan\textsuperscript{\rm 2,\rm 6},
    Zhijie Zhong\textsuperscript{\rm 2,\rm 1},
    Yuchen Fan\textsuperscript{\rm 2,\rm 1},
    Biqing Qi\textsuperscript{\rm 2}\corresponding,
    Bowen Zhou\textsuperscript{\rm 2}
}
\affiliations{
    \textsuperscript{\rm 1}Shanghai Jiao Tong University,
    \textsuperscript{\rm 2}Shanghai Artificial Intelligence Laboratory,
    \textsuperscript{\rm 3}Tsinghua University,\\
    \textsuperscript{\rm 4}Zhejiang University,
    \textsuperscript{\rm 5}Harbin Institute of Technology,
    \textsuperscript{\rm 6}University of Science and Technology of China
}

\begin{document}

\maketitle

\begin{abstract}
Long-context reasoning remains a critical bottleneck for large language models, as recent recurrent-memory approaches face two inherent challenges: sequential chunk-wise updates can overwrite early critical evidence with later irrelevant content, and serial inter-chunk dependencies limit parallelism and cause latency to increase with context length.
To address these issues, we propose \textbf{PI-Mem} (Parallel-Iterative Memory), a mechanism that processes all chunks in parallel and iteratively refines a shared memory over a bounded number of turns.
In each turn, PI-Mem reads all chunks in parallel conditioned on the current memory, selects new or complementary evidence from each chunk, and merges the selected evidence into a compact shared memory for the next turn.
To discourage redundant turns, we optimize the workflow through reinforcement learning with an auxiliary turn-efficiency reward, enabling the model to adaptively exit once sufficient evidence has been accumulated.
We evaluate PI-Mem with Qwen3.5-35B-A3B and Qwen2.5-7B on the HotpotQA benchmark across context lengths up to 3.6 million tokens and find that it outperforms the recurrent-memory baseline by +6.25 and +7.81 absolute points while achieving 6.1$\times$ and 2.1$\times$ inference speedups, respectively.
These results demonstrate that PI-Mem breaks the accuracy--efficiency trade-off in long-context reasoning and provides a scalable approach to complex multi-hop question answering over extremely long documents.
\end{abstract}

\begin{links}
    \link{Code}{https://github.com/JetAstra/PI-Mem}
\end{links}

\section{Introduction}

The ability to reason over long contexts is increasingly important for large language models (LLMs), supporting applications such as long-document question answering~\citep{geminiteam2025gemini25,yang2025qwen251m}, multi-turn dialogue~\citep{maharana2024locomo,li2024helloagain,wu2025longmemeval}, repository-level code understanding~\citep{yang2024sweagent,hui2024qwen25coder}, and agentic workflows over large external contexts~\citep{yao2023react,agashe2024agents,kimiteam2026kimik25visualagentic}.
Yet model performance often degrades as context length grows, highlighting the persistent difficulty of identifying, retaining, and effectively utilizing relevant information across distant parts of the context~\citep{hsieh2024ruler,du2025contextlengthhurtsllm}.

\begin{figure}[t]
\centering
\includegraphics[width=\columnwidth]{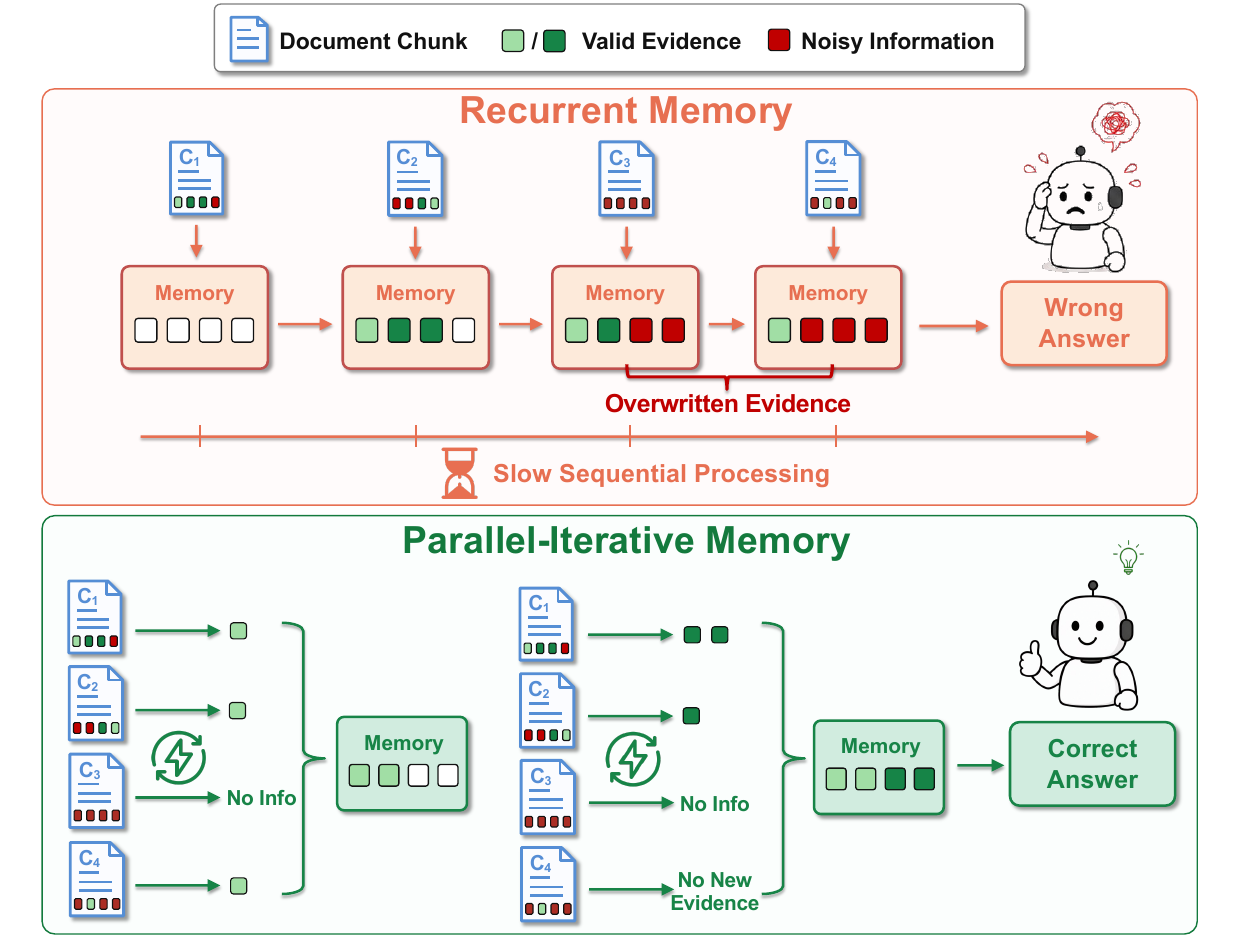}
\caption{Top: Recurrent-memory workflows process chunks sequentially and repeatedly rewrite a running memory state, potentially overwriting early evidence and creating strict serial dependencies. Bottom: PI-Mem reads chunks in parallel conditioned on a shared global memory, improving evidence preservation and reducing inference latency.}
\label{fig:teaser}
\end{figure}

One family of approaches addresses this challenge by extending the natively supported context window through positional-encoding interpolation or extrapolation, thereby enabling direct processing of longer sequences without discarding input context~\citep{chen2023extendingcontextwindow,peng2026yarnefficientcontextwindow,liu2025comprehensivesurvey}.
In practice, such extensions are generally reliable only within a limited extrapolation range and still incur the high computational cost of dense attention at ultra-long context lengths.
To address this efficiency bottleneck, sparse attention restricts computation to selected tokens or key-value blocks, whereas recurrent formulations of linear attention summarize preceding tokens in a compact state.
Nevertheless, these architectures are costly to train and often rely on high-quality synthetic long-context corpora~\citep{yuan2025nativesparseattention,lu2025moba,yang2025gateddeltanetworks,qwen3.5}, whose construction and curation remain difficult at scale across diverse reasoning tasks.

These limitations motivate recurrent-memory processing as a complementary strategy for handling long inputs.
A recurrent-memory mechanism maintains a fixed-length textual memory as a compact state throughout input processing.
At each step, the model processes the current chunk together with the previous memory and generates an updated memory that overwrites the previous state.
Once all chunks have been processed, the model generates the final answer conditioned on the resulting memory~\citep{yu2025memagent,sheng2026memorizestopgatedrecurrent,shi2025lookbackreasonforward}.
In principle, bounding the context of each memory update allows the workflow to process inputs of arbitrary length at a computational cost that grows linearly rather than quadratically with input length.
However, as illustrated in Figure~\ref{fig:teaser}, this recurrent-memory design faces two challenges.
First, successive memory updates compress each new chunk and the existing fixed-length memory into a new state, potentially overwriting early evidence with later noisy information before the relevance of the early evidence becomes apparent and thereby impairing evidence preservation and integration across distant chunks.
Second, the recurrent-memory workflow imposes a strict inter-chunk dependency, forcing each chunk to await the preceding memory update and increasing inference latency with context length.

To address these challenges, we introduce {PI-Mem}, a {P}arallel-{I}terative Memory mechanism for long-context reasoning.
PI-Mem gathers evidence from all chunks in \emph{parallel} and uses it to update a shared memory over a bounded number of turns.
Specifically, we define each memory-update \emph{turn} as a complete \emph{read-select-merge} cycle.
The read step processes all chunks in parallel, with each chunk-level read conditioned on the same global memory, thereby avoiding sequential memory overwrites that may discard earlier evidence and reducing inference latency through parallel chunk processing.
PI-Mem then selects observations that provide new or complementary evidence and merges them with the current memory to produce an updated global memory.
In the next turn, all chunks are read again conditioned on the updated memory, allowing evidence discovered in one chunk to guide extraction from other chunks and thereby enabling cross-chunk information exchange.
This cycle repeats until either no chunk yields useful new evidence or a predefined maximum number of turns is reached, after which PI-Mem generates the final answer from the question and the consolidated global memory.
To train PI-Mem, we optimize the entire workflow end-to-end with reinforcement learning (RL), combining an answer-accuracy reward with a turn-efficiency reward that discourages redundant turns.

Experiments with Qwen3.5-35B-A3B and Qwen2.5-7B demonstrate the effectiveness and efficiency of PI-Mem. On HotpotQA (HQA) at 3.6M tokens, PI-Mem improves over MemAgent by +6.25 and +7.81 absolute points and delivers inference speedups of 6.1$\times$ and 2.1$\times$, respectively.

Our main contributions are summarized as follows:
\begin{itemize}
    \item We propose PI-Mem, a parallel-iterative memory mechanism for long-context reasoning that reads all chunks in parallel conditioned on a shared memory and iteratively refines the memory over a bounded number of turns.
    \item We optimize the workflow end-to-end with RL to improve long-context performance and introduce a turn-efficiency reward that discourages redundant turns.
    \item We empirically show that PI-Mem outperforms the recurrent-memory baseline and substantially reduces inference latency on HotpotQA at context lengths up to 3.6M tokens, while also improving performance across diverse long-context benchmarks.
\end{itemize}

\section{Related Work}

\paragraph{Memory-Based Context Management.}
Memory-based approaches retain salient information in external stores or compact textual states to operate beyond limited context windows.
Early work introduced hierarchical virtual context management, while production-oriented layers extract, consolidate, and retrieve conversational information~\citep{packer2023memgpt,chhikara2025mem0}.
Recent workflows recurrently update compact states over document chunks or interaction steps; related variants reconstruct evolving reports or reason over recalled compressed memories~\citep{yu2025memagent,zhou2025mem1,chen2026iterresearchrethinkinglonghorizonagents,chen2026dynamiclongcontextreasoning}.
These designs bound the active context but retain sequential state transitions.
Later work improves selectivity through update and exit gates, learned memory construction and editing, active retrieval and writing, multi-scale folding, and adaptive routing~\citep{sheng2026memorizestopgatedrecurrent,wang2025memalpha,zhang2025memoryaction,wang2026infmem,sun2025contextfolding,ye2025agentfold,feng2026agentswing}.
Our approach instead conditions all chunk-level reads within each turn on the same memory, improving evidence preservation and reducing inference latency.

\paragraph{Reinforcement Learning for Long-Context Reasoning.}
RL with verifiable rewards, together with group-relative policy optimization and its variants, has substantially improved multi-step reasoning~\citep{shao2024deepseekmathpushinglimitsmathematical,yu2025dapo}.
Long-context studies extend this paradigm through supervised warm-up, progressive context curricula, difficulty-aware sampling, self-play verification, and synthetic multi-hop tasks with distractors~\citep{wan2025qwenlongl1,yang2025spell,wang2025loongrl}.
These strategies stabilize optimization, adapt task difficulty as training progresses, and provide scalable reward signals when annotated long-context data are scarce.
Beyond outcome-only supervision, dense, verifiable context rewards directly guide grounding and evidence selection~\citep{chen2026longrlvr}.
RL has also been used to train memory-augmented workflows, including iterative memory fusion, historical-memory retrieval, and policies that control retrieval, writing, and stopping through final-answer and step-level signals~\citep{shen2025qwenlongl15,shi2025lookbackreasonforward,wang2026infmem}.
Our approach applies trajectory-level RL to a parallel-iterative workflow, training the policy to coordinate evidence selection and memory refinement and discouraging redundant turns.

\begin{figure*}[t]
\centering
\includegraphics[width=\textwidth]{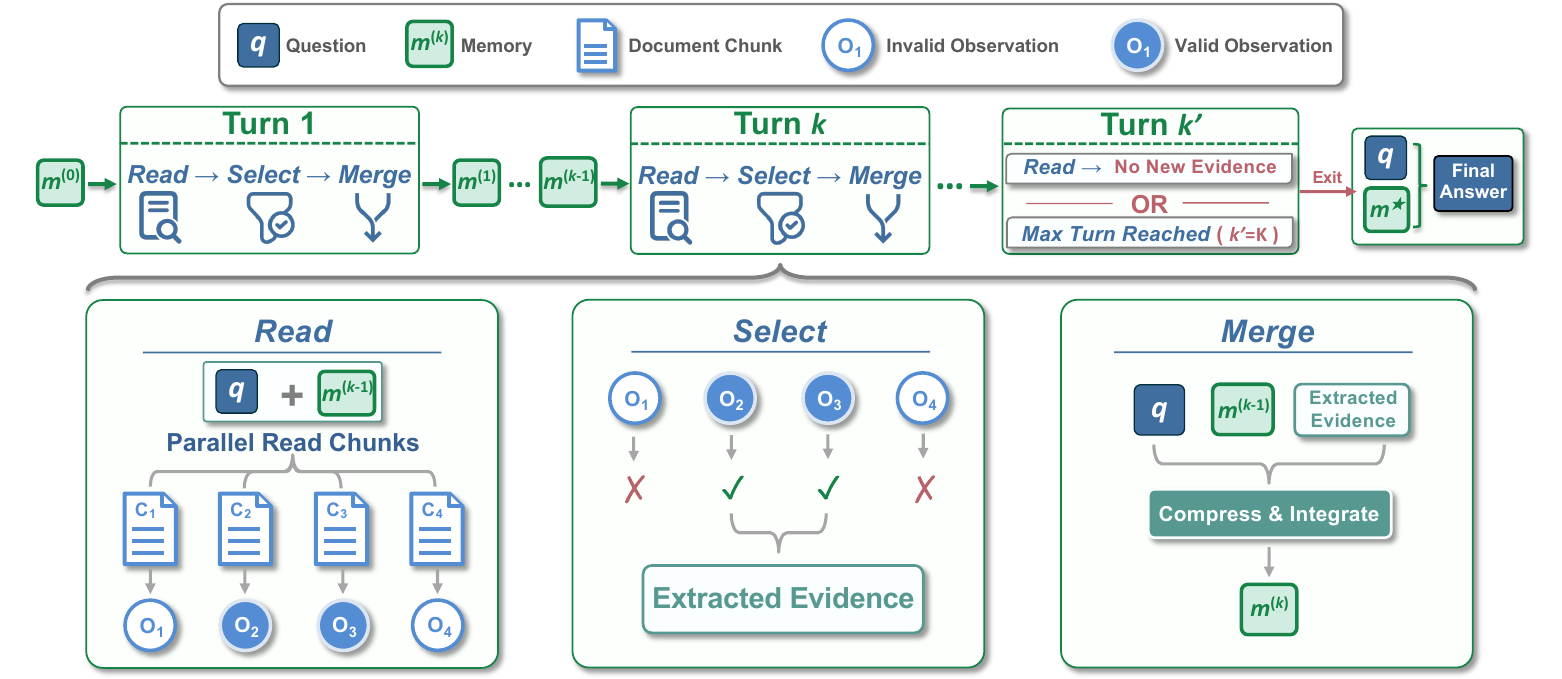}
\caption{Overview of the PI-Mem workflow. PI-Mem reads all chunks in parallel conditioned on a shared global memory and iteratively refines this memory through read-select-merge turns.}
\label{fig:method}
\end{figure*}

\section{Method}

In this section, we first describe the PI-Mem inference workflow and then present its RL training procedure.

\subsection{PI-Mem Inference Workflow}

Figure~\ref{fig:method} provides an overview of the PI-Mem inference workflow, and Algorithm~\ref{alg:algorithm} presents its procedural details.
PI-Mem reads all chunks in parallel and iteratively refines a shared memory over a bounded number of turns.
We define each memory-update \emph{turn} as a complete \emph{read-select-merge} cycle: PI-Mem reads all chunks conditioned on the current global memory, selects observations with new or complementary evidence, and merges them into an updated global memory.
We first detail the three operations within each turn and then describe iterative refinement and adaptive exit across turns.

\begin{algorithm}[tb]
\caption{PI-Mem Inference Workflow}
\label{alg:algorithm}
\textbf{Input}: Question $q$, context $C$, model $\pi_\theta$, maximum number of turns $K$ \\
\textbf{Output}: Final answer $a$ \\
\textbf{Initialize}: Split $C$ into chunks $\{c_i\}_{i=1}^n$; \\
\phantom{\textbf{Initialize}:} $m^{(0)} \leftarrow \texttt{EMPTY}$
\begin{algorithmic}[1] 
\STATE $m^\star \leftarrow m^{(0)}$
\FOR{$k=1$ to $K$}
    \STATE Read each chunk $c_i$ in \emph{parallel}:
    \STATE \hspace{1em} $o_i^{(k)} \leftarrow \textsc{ReadCall}(\pi_\theta, q, c_i, m^{(k-1)})$
    \STATE $O^{(k)} \leftarrow \{\,o_i^{(k)} \mid \chi(o_i^{(k)}) = \text{\texttt{yes}}\,\}$
    \IF{$O^{(k)} = \emptyset$}
        \STATE \textbf{break}
    \ELSE
        \STATE $m^{(k)} \leftarrow \textsc{MergeCall}(\pi_\theta, q, m^{(k-1)}, O^{(k)})$
        \STATE $m^\star \leftarrow m^{(k)}$
    \ENDIF
\ENDFOR
\STATE $a \leftarrow \textsc{FinalCall}(\pi_\theta, q, m^\star)$
\STATE \textbf{return} $a$
\end{algorithmic}
\end{algorithm}

\paragraph{Read-Select-Merge Turn.}
For a general QA task, we denote the question by $q$, the full background context by $C$, and the model by $\pi_\theta$. We first split $C$ into equal-sized chunks $\{c_i\}_{i=1}^n$ and initialize the global memory $m^{(0)}$ as an empty memory. At turn $k$, PI-Mem updates the global memory through a \emph{read-select-merge} turn:
\begin{equation}
    m^{(k)} = \textsc{Turn}(\pi_\theta, q, m^{(k-1)}, \{c_i\}_{i=1}^n).
\end{equation}
The goal of each turn is to update the global memory by processing all chunks in parallel.

In the \textbf{read} step, PI-Mem processes all chunks in parallel.
For each chunk $c_i$, \textsc{ReadCall} places $q$, $c_i$, and $m^{(k-1)}$ into a chunk-reading template and calls $\pi_\theta$ to generate a chunk-level observation $o_i^{(k)}$.
Rather than successively overwriting the memory as in sequential recurrent processing, PI-Mem reads all chunks independently conditioned on the same memory $m^{(k-1)}$, thereby reducing the risk that evidence from early chunks is overwritten by noisy information from later chunks.
This independence also enables PI-Mem to read chunks in parallel and efficiently batch the corresponding calls, allowing ultra-long sequences to be processed with lower inference latency.
Each chunk call must explicitly output a check signal, either \texttt{\textless check\textgreater yes\textless/check\textgreater} or \texttt{\textless check\textgreater no\textless/check\textgreater}.
The model outputs \texttt{yes} when $c_i$ provides evidence relevant to $q$ that is new or complementary to $m^{(k-1)}$.
When the check signal is \texttt{yes}, the output includes the corresponding evidence needed for answering $q$; otherwise, the model outputs only \texttt{no}.

The \textbf{select} step filters observations by the \texttt{\textless check\textgreater} signal. Let $\chi(o_i^{(k)})$ denote the value enclosed by the check tags; the selected observation set is defined as
\begin{equation}
    O^{(k)} = \{\,o_i^{(k)} \mid \chi(o_i^{(k)}) = \text{\texttt{yes}}\,\}.
\end{equation}
This filtering removes observations that provide no useful update, so the memory update focuses on new or complementary evidence rather than irrelevant or redundant text.

The \textbf{merge} step consolidates the selected observations into the updated global memory. Specifically, \textsc{MergeCall} places $q$, $m^{(k-1)}$, and $O^{(k)}$ into a merge template and invokes $\pi_\theta$ to produce $m^{(k)}$. A straightforward alternative is to concatenate all positive chunk outputs, but this can still produce an overly long intermediate context when many chunks contain relevant evidence. The merge step instead compresses and integrates evidence from different chunks, removes redundancy, preserves details necessary for answering $q$, and prevents unbounded memory growth as the number of selected observations increases. Thus, a single turn updates the global memory through parallel evidence extraction, selective filtering, and compact cross-source integration.

\paragraph{Iterative Refinement and Exit Mechanism.}
After turn $k$ finishes, PI-Mem feeds the merged memory $m^{(k)}$ into the next read step and repeats the read-select-merge cycle. Each later turn therefore reads all chunks again, conditioned on evidence discovered in previous turns. Compared with a single greedy turn, in which chunk relevance is judged only from the initial empty memory, iterative refinement reduces the risk of overlooking evidence whose importance becomes clear only after other facts have been identified and enables cross-chunk information exchange through the shared memory.

PI-Mem further uses the check signal as an \textbf{adaptive exit mechanism}. The loop automatically stops when no chunk produces useful new evidence, i.e., $O^{(k)}=\emptyset$, or when it reaches the maximum number of turns $K$. Finally, PI-Mem invokes the model once more to answer using only the question $q$ and the final global memory $m^\star$, rather than the original long context, so that final inference remains compact and grounded in the consolidated evidence.

\begin{table*}[t]
\centering
{
\begin{tabular}{l|cccccccccc|c}
\toprule
Method & 7K & 14K & 28K & 56K & 112K & 224K & 448K & 896K & 1.8M & 3.6M & Avg. \\
\midrule
\multicolumn{12}{c}{\textit{\textbf{Qwen3.5-35B-A3B}}} \\
\midrule
Vanilla & \textbf{87.50} & 79.69 & 82.81 & 81.25 & 78.12 & 73.44 & 62.50 & 37.50 & 34.38 & 29.69 & 64.69 \\
YaRN & 82.81 & 81.25 & 79.69 & \textbf{82.81} & 78.12 & 68.75 & 59.38 & 56.25 & 29.69 & 14.06 & 63.28 \\
RAG & 84.38 & \textbf{82.81} & 71.88 & 67.19 & 46.88 & 54.69 & 51.56 & 53.12 & 46.88 & 43.75 & 60.31 \\
MemAgent & 84.38 & 81.25 & 82.81 & 81.25 & 79.69 & 68.75 & 73.44 & 71.88 & 71.88 & 70.31 & 76.56 \\
\rowcolor{controlled}\textbf{PI-Mem} & \textbf{87.50} & \textbf{82.81} & \textbf{84.38} & \textbf{82.81} & \textbf{84.38} & \textbf{79.69} & \textbf{82.81} & \textbf{76.56} & \textbf{75.00} & \textbf{76.56} & \textbf{81.25} \\
\midrule
\multicolumn{12}{c}{\textit{\textbf{Qwen2.5-7B}}} \\
\midrule
Vanilla & 60.94 & 54.69 & 50.00 & 23.44 & 0.00 & 0.00 & 0.00 & 0.00 & 0.00 & 0.00 & 18.91 \\
YaRN & 56.25 & 62.50 & 62.50 & 45.31 & 34.38 & 0.00 & 0.00 & 0.00 & 0.00 & 0.00 & 26.09 \\
RAG & 57.81 & 46.88 & 46.88 & 45.31 & 39.06 & 42.19 & 29.69 & 43.75 & 26.56 & 32.81 & 41.09 \\
MemAgent & 81.25 & 81.25 & 75.00 & \textbf{82.81} & 76.56 & 75.00 & 76.56 & 75.00 & 78.12 & 73.44 & 77.50 \\
GRU-Mem & 81.25 & 79.69 & 77.34 & 76.56 & 74.22 & 73.44 & 71.88 & 76.56 & N/A & N/A & 76.37 \\
ReMemR1 & 82.3 & 82.8 & 81.1 & 78.9 & 82.0 & 79.7 & 80.0 & 80.8 & N/A & N/A & 80.95 \\
\rowcolor{controlled}\textbf{PI-Mem} & \textbf{82.81} & \textbf{82.81} & \textbf{81.25} & \textbf{82.81} & \textbf{89.06} & \textbf{85.94} & \textbf{82.81} & \textbf{84.38} & \textbf{87.50} & \textbf{81.25} & \textbf{84.06} \\
\bottomrule
\end{tabular}
}
\caption{Length-grouped RULER HQA results. Scores are reported across context lengths from 7K to 3.6M tokens. The averages for GRU-Mem and ReMemR1 use the eight reported lengths.}
\label{tab:ruler_hqa_length_results}
\end{table*}

\subsection{Trajectory-Level Reinforcement Learning}

We adopt GRPO~\citep{shao2024deepseekmathpushinglimitsmathematical} to optimize the multi-call workflow end to end. During RL training, each rollout for a query defines a complete trajectory. Since the workflow invokes the model multiple times, each trajectory contains multiple model-call samples, including read, merge, and final-answer calls. We denote the $i$-th trajectory by
\begin{equation}
\label{eq:rl-trajectory}
\tau_i = \{(x_{i,s}, y_{i,s})\}_{s=1}^{S_i},
\end{equation}
where $S_i$ is the number of model-call samples in trajectory $\tau_i$, $s$ indexes one such model call, $x_{i,s}$ is the corresponding phase-specific prompt constructed from the workflow state, and $y_{i,s}$ is the generated response. For each training query, we sample a group of $G$ complete trajectories $\{\tau_i\}_{i=1}^G$ from the old policy $\pi_{\theta_{\mathrm{old}}}$ and use all model-call samples in these trajectories for policy optimization. We next describe how the trajectory-level reward is constructed and used for end-to-end workflow optimization.

\paragraph{Reward Design.}
We assign one reward to each completed trajectory $\tau_i$ based on its final answer $a_i$ and turn count $k_i$. Let $r_{\mathrm{acc}}(a_i)$ denote the accuracy reward and $K$ the maximum number of turns. We augment answer accuracy with a turn-efficiency bonus:
\begin{equation}
\label{eq:final-reward}
r_{\mathrm{turn}}(\tau_i) = \frac{K-k_i}{K-1};
\ 
R_i = r_{\mathrm{acc}}(a_i) + \lambda_{\mathrm{turn}} r_{\mathrm{turn}}(\tau_i).
\end{equation}
The turn bonus assigns higher rewards to trajectories that exit in fewer turns, and $\lambda_{\mathrm{turn}}$ controls its contribution. It discourages redundant refinement once the memory is sufficient and encourages the workflow to exit when no new evidence is found, reducing latency.

\paragraph{Trajectory-Level Optimization.}
This scalar reward is then broadcast to every model-call sample $(x_{i,s},y_{i,s})$ in the same trajectory.
Following Dr. GRPO~\citep{liu2025understandingr1zero}, we omit normalization by the group standard deviation and compute the group-relative advantage over complete trajectories rather than individual calls:
\begin{equation}
\label{eq:rl-advantage}
A_i = R_i - \frac{1}{G}\sum_{j=1}^{G} R_j.
\end{equation}
Given this trajectory-level advantage, each token in every model-call sample of trajectory $\tau_i$ uses the same $A_i$ in the per-token clipped surrogate term:
\begin{equation}
\label{eq:rl-token-surrogate}
\begin{aligned}
\ell_{i,s,t}(\theta)
= &\min\!\Big(\rho_{i,s,t}(\theta)A_i, \\
&\operatorname{clip}(\rho_{i,s,t}(\theta),1-\varepsilon,1+\varepsilon)A_i\Big),
\end{aligned}
\end{equation}
where $\varepsilon$ denotes the clipping ratio, and the token-level policy ratio is
\begin{equation}
\label{eq:rl-token-ratio}
\rho_{i,s,t}(\theta)
= \frac{\pi_\theta(y_{i,s,t}\mid x_{i,s},y_{i,s,<t})}
        {\pi_{\theta_{\mathrm{old}}}(y_{i,s,t}\mid x_{i,s},y_{i,s,<t})}.
\end{equation}
The final objective uses DAPO-style token-level aggregation~\cite{yu2025dapo} over all model-call samples in the sampled trajectories. Let $Z=\sum_{i=1}^{G}\sum_{s=1}^{S_i}|y_{i,s}|$ denote the total number of generated tokens. The objective is
\begin{equation}
\label{eq:rl-objective}
\begin{aligned}
\mathcal{J}(\theta)
&=
\mathbb{E}_{\substack{
(q,a)\sim\mathcal{D},
\{y_{i,s}\}{\sim}
\pi_{\theta_{\mathrm{old}}}(\cdot\mid x_{i,s})
}}\\
\Bigg[
&
\frac{1}{Z}\sum_{i=1}^{G}\sum_{s=1}^{S_i}
\sum_{t=1}^{|y_{i,s}|}
\Big(
\ell_{i,s,t}(\theta)
-\beta \mathrm{KL}
(
\pi_\theta
\|\,
\pi_{\mathrm{ref}}
)
\Big)
\Bigg].
\end{aligned}
\end{equation}
Here, $(q,a)$ denotes a question and its reference answer sampled from the data distribution $\mathcal{D}$. The coefficient $\beta$ scales the token-level KL divergence between the current policy $\pi_\theta$ and the reference policy $\pi_{\mathrm{ref}}$.

\section{Experiments}

\subsection{Training Setup}

We evaluate our method on two representative models: Qwen3.5-35B-A3B~\citep{qwen3.5}, an MoE model with hybrid attention, and Qwen2.5-7B-Instruct~\citep{qwen2025qwen25technicalreport}, a dense model with full attention.

For the Qwen3.5-35B-A3B experiments, following the data construction protocol of MemAgent~\citep{yu2025memagent}, we synthesize long-context QA training samples by embedding HotpotQA~\citep{yang2018hotpotqadatasetdiverseexplainable} gold paragraphs into distractor documents sampled from the same dataset. Each training sample contains 1,000 documents, yielding a context length of approximately 140K tokens.
PI-Mem splits the context into 15K-token chunks.
Across the workflow, all generation stages use the same maximum output length of 4,096 tokens, including chunk-level memory generation, merged-memory generation, and final-answer generation.
Thinking mode is disabled during training to improve rollout efficiency and reduce the generated output length of each model call.
We set the maximum number of turns to $K=3$.
For a fair comparison, the MemAgent baseline is trained with the same RL hyperparameters: a rollout batch size of 128 prompts, a GRPO group size of 8 rollouts per prompt, and 80 rollout steps, corresponding to 10,240 training samples counted at the prompt level.

For Qwen2.5-7B, each HotpotQA training sample contains 200 documents (approximately 28K tokens). PI-Mem uses 5K-token chunks and a maximum output length of 1,024 tokens. We use 16 rollouts per prompt and 240 rollout steps, and compare against the officially released MemAgent checkpoint with the same backbone.

\subsection{Evaluation Setup}

\paragraph{Baselines.}
We compare PI-Mem with direct-inference, retrieval-augmented, and recurrent-memory baselines. Vanilla denotes standard direct inference using the backbone model. We also include YaRN~\citep{peng2026yarnefficientcontextwindow}, a commonly used positional-encoding extrapolation method, as a direct-inference baseline. RAG~\citep{lewis2020retrievalaugmented} retrieves relevant chunks from the long context and performs direct inference over the retrieved evidence. We include MemAgent~\citep{yu2025memagent}, GRU-Mem~\citep{sheng2026memorizestopgatedrecurrent}, and ReMemR1~\citep{shi2025lookbackreasonforward} as recurrent-memory workflow baselines. We report the \textbf{RL-trained variants} of MemAgent and PI-Mem and use the published HQA results for GRU-Mem and ReMemR1.

\paragraph{Benchmarks.}
We evaluate long-context performance on RULER~\citep{hsieh2024ruler} and LongBench v2~\citep{bai2024longbench2}. RULER is a synthetic benchmark that probes retrieval, multi-hop tracing, aggregation, and question answering over long contexts. LongBench v2 contains 503 challenging multiple-choice questions across six realistic task categories. In our RULER results, all tasks \emph{except HQA} are treated as out-of-distribution (OOD) tasks, while HQA is considered in-distribution because the RL training data is synthesized from HQA.
We report LongBench v2 results only for Qwen3.5, as all evaluated methods on Qwen2.5 performed at approximately the random-guessing level on this four-choice benchmark in our experiments, making method-level comparisons difficult to interpret.

\paragraph{Implementation Details.}
Thinking mode is disabled for Qwen3.5 and is not applicable to Qwen2.5. We use temperature $=0.7$ and top-$p=0.95$, and set the maximum output length to 4,096 tokens for Qwen3.5 and 1,024 tokens for Qwen2.5.
We use 8 NVIDIA H200 GPUs with tensor parallelism set to 2 for inference. 
For RULER, each task at each context length is evaluated with 64 samples.

\subsection{Main Results}

\begin{table*}[t]
\centering
{
\setlength{\tabcolsep}{9pt}
\begin{tabular}{l|cccccccc|c}
\toprule
Method & 8K & 16K & 32K & 64K & 128K & 256K & 512K & 1M & Avg. \\
\midrule
\multicolumn{10}{c}{\textit{\textbf{Qwen3.5-35B-A3B}}} \\
\midrule
Vanilla & 98.67 & 98.39 & 97.64 & 97.96 & 97.40 & 96.08 & 95.81 & 75.56 & 94.69 \\
YaRN & 97.92 & 96.59 & 96.74 & 97.02 & 97.46 & 97.22 & 96.19 & 86.58 & 95.72 \\
RAG & 98.31 & 93.07 & 94.32 & 95.18 & 93.25 & 87.02 & 77.90 & 75.24 & 89.29 \\
MemAgent & 94.55 & 92.92 & 95.18 & 94.05 & 92.46 & 90.84 & 86.42 & 83.86 & 91.28 \\
\rowcolor{controlled}\textbf{PI-Mem} & \textbf{98.92} & \textbf{98.65} & \textbf{98.06} & \textbf{98.42} & \textbf{97.69} & \textbf{98.38} & \textbf{97.30} & \textbf{96.88} & \textbf{98.04} \\
\midrule
\multicolumn{10}{c}{\textit{\textbf{Qwen2.5-7B}}} \\
\midrule
Vanilla & 86.57 & 87.17 & 84.48 & 60.65 & 22.72 & 0.00 & 0.00 & 0.00 & 42.70 \\
YaRN & 84.24 & 83.03 & 78.82 & 72.48 & 57.13 & 0.00 & 0.00 & 0.00 & 46.96 \\
RAG & 80.58 & 79.30 & 77.57 & 76.49 & 74.07 & 68.05 & 63.45 & 58.70 & 72.28 \\
MemAgent & 91.66 & 90.33 & 89.31 & 87.73 & 83.73 & 80.30 & 79.51 & 74.83 & 84.68 \\
\rowcolor{controlled}\textbf{PI-Mem} & \textbf{91.72} & \textbf{91.78} & \textbf{92.10} & \textbf{92.49} & \textbf{91.75} & \textbf{91.38} & \textbf{89.14} & \textbf{88.39} & \textbf{91.09} \\
\bottomrule
\end{tabular}
}
\caption{Length-grouped RULER out-of-distribution results. Scores are reported across context lengths from 8K to 1M tokens.}
\label{tab:ruler_ood_length_results}
\end{table*}

Tables~\ref{tab:ruler_hqa_length_results} and~\ref{tab:ruler_ood_length_results} summarize the length-grouped RULER HQA and OOD results. On HQA, PI-Mem achieves the best average score for both backbones. With Qwen3.5-35B-A3B, it improves the average from 76.56 for MemAgent to 81.25 and remains strong at the longest 3.6M-token setting, where Vanilla and YaRN degrade sharply. With Qwen2.5-7B, PI-Mem improves the average from 77.50 to 84.06 and matches or outperforms MemAgent across all reported lengths, with especially large gains in the longer-context regime.

On RULER OOD tasks, PI-Mem also shows stronger robustness as context length increases. For Qwen3.5-35B-A3B, it obtains the best average score of 98.04, outperforming MemAgent by 6.76 points and surpassing both direct-inference baselines. At 1M tokens, PI-Mem reaches 96.88 while MemAgent drops to 83.86. For Qwen2.5-7B, PI-Mem improves the average from 84.68 to 91.09 and raises the 1M-token score from 74.83 to 88.39. Together, these results show that the parallel-iterative workflow remains robust as context length grows, with gains that persist across both backbones and extend beyond the HQA-based training distribution.

\begin{table}[t]
\centering
{\small
\begin{tabular}{l|c|cc|ccc}
\toprule
\multirow{2}{*}{\textbf{Method}} &
\multirow{2}{*}{\textbf{All}} &
\multicolumn{2}{c|}{\textbf{Difficulty}} &
\multicolumn{3}{c}{\textbf{Length}} \\
& & Easy & Hard & Short & Med. & Long \\
\midrule
Vanilla & 50.7 & 55.7 & 47.6 & 56.1 & 47.9 & 47.2 \\
YaRN & 51.5 & 52.1 & \textbf{51.1} & \textbf{57.8} & 48.4 & 47.2 \\
RAG & 48.9 & 53.6 & 46.0 & 51.7 & 44.7 & 52.8 \\
MemAgent& 52.3 & 59.9 & 47.6 & 55.6 & 48.8 & 53.7 \\
\rowcolor{controlled}\textbf{PI-Mem} & \textbf{54.1} & \textbf{66.1} & 46.6 & 53.3 & \textbf{49.8} & \textbf{63.9} \\
\bottomrule
\end{tabular}
}
\caption{LongBench v2 results. Following the grouping defined by the benchmark, we report results by difficulty (Easy and Hard) and context length (Short, Medium, and Long).}
\label{tab:longbenchv2_results}
\end{table}

Table~\ref{tab:longbenchv2_results} reports the LongBench v2 results. PI-Mem achieves the best overall score of 54.1, outperforming MemAgent by 1.8 points and the vanilla baseline by 3.4 points. These results show that the proposed workflow improves long-context performance across realistic tasks.

\begin{table}[t]
\centering
{\footnotesize
\setlength{\tabcolsep}{3.5pt}
\begin{tabular}{l|ccc|ccc}
\toprule
\multirow{2}{*}{\textbf{Method}} &
\multicolumn{3}{c|}{\textit{\textbf{Qwen3.5-35B-A3B}}} &
\multicolumn{3}{c}{\textit{\textbf{Qwen2.5-7B}}} \\
& 256K & 512K & 1M & 256K & 512K & 1M \\
\midrule
{MemAgent} &
70.3\% & 74.2\% & 58.6\% & 73.8\% & 75.8\% & 74.6\% \\
{\textbf{PI-Mem}} &
100\% & 100\% & 100\% & 98.8\% & 100\% & 100\% \\
\bottomrule
\end{tabular}
}
\caption{Final-memory evidence coverage on RULER MV-NIAH, measured as the proportion of ground-truth values present in the final memory.}
\label{tab:mv_memory_analysis}
\end{table}

\begin{figure}[t]
\centering
\includegraphics[width=\columnwidth]{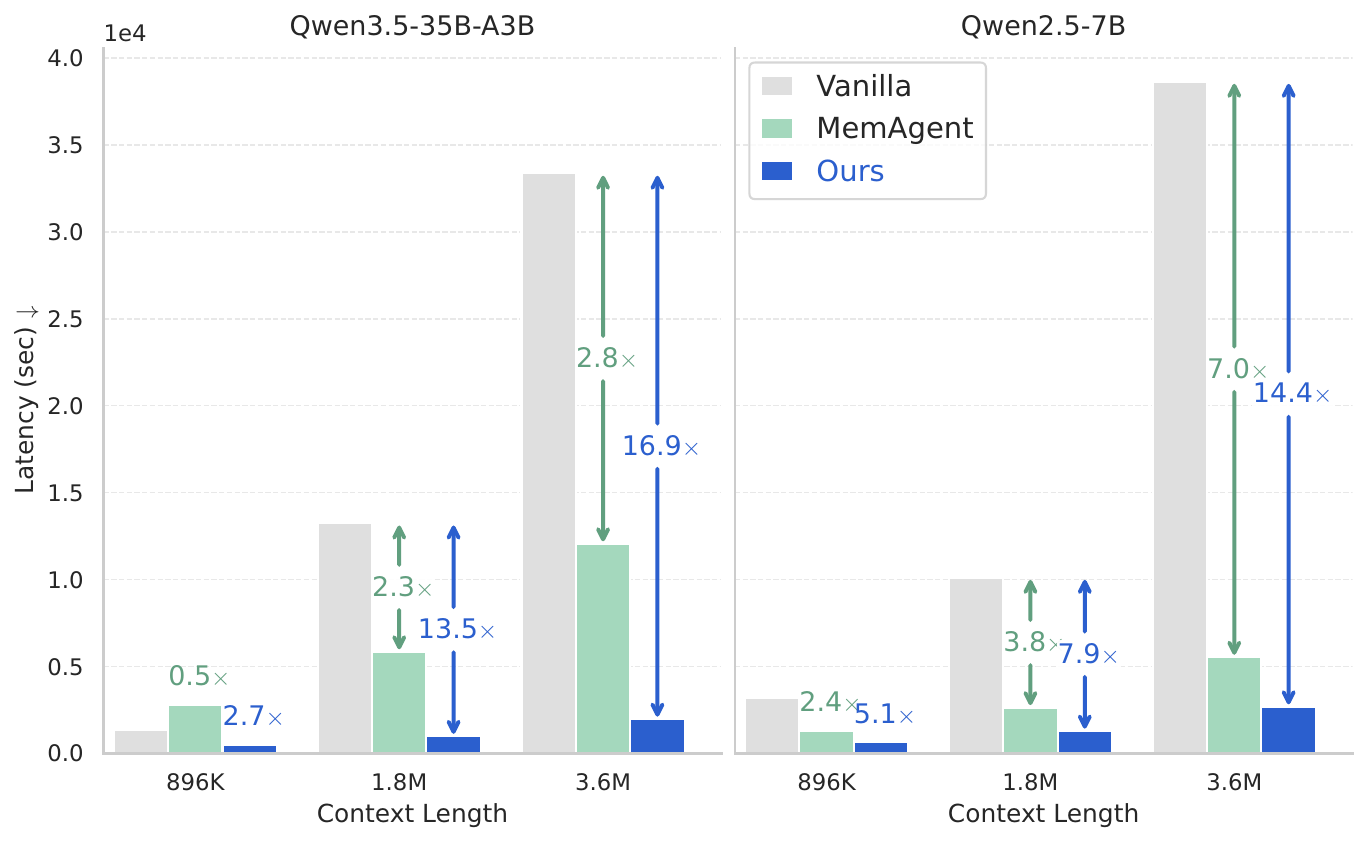}
\caption{End-to-end HQA inference latency at ultra-long context lengths with Qwen3.5-35B-A3B and Qwen2.5-7B. Reported speedups are relative to Vanilla.}
\label{fig:latency}
\end{figure}

\paragraph{Evidence Retention Analysis.}
To directly assess whether recurrent memory updates overwrite earlier evidence, we measure final-memory evidence coverage on RULER Multi-values Needle-in-a-Haystack (MV-NIAH), where coverage is defined as the proportion of ground-truth values present in the final memory~\citep{hsieh2024ruler}. The task requires retrieving four independent values associated with the same key from positions distributed across a long context and thus provides a controlled diagnostic of evidence retention. Table~\ref{tab:mv_memory_analysis} shows a clear contrast: PI-Mem retains almost all target evidence in the final memory, whereas MemAgent loses a substantial portion of that evidence over successive recurrent updates. This comparison indicates that parallel reading preserves the collected evidence more effectively by mitigating the overwrite caused by recurrent memory updates.

\paragraph{Latency Analysis.}
Figure~\ref{fig:latency} compares end-to-end inference latency on ultra-long HQA inputs. At 3.6M tokens, PI-Mem is 16.9$\times$ faster than Vanilla and 6.1$\times$ faster than MemAgent with Qwen3.5-35B-A3B; the corresponding speedups with Qwen2.5-7B are 14.4$\times$ and 2.1$\times$. Vanilla keeps the full context active during inference, incurring substantial full-context prefill and KV-cache costs. MemAgent bounds the context of each call through chunking, but its recurrent updates form a serial critical path that prevents concurrent processing of chunks from the same sequence. PI-Mem removes this dependency by executing chunk-level \textsc{ReadCall} operations in parallel within each turn and batching the corresponding calls across examples, thereby improving GPU utilization and reducing latency.

\subsection{Ablation Study}

For computational efficiency, we conduct all ablations in this section with Qwen2.5-7B-Instruct.

\begin{figure}[t]
\centering
\includegraphics[width=\columnwidth]{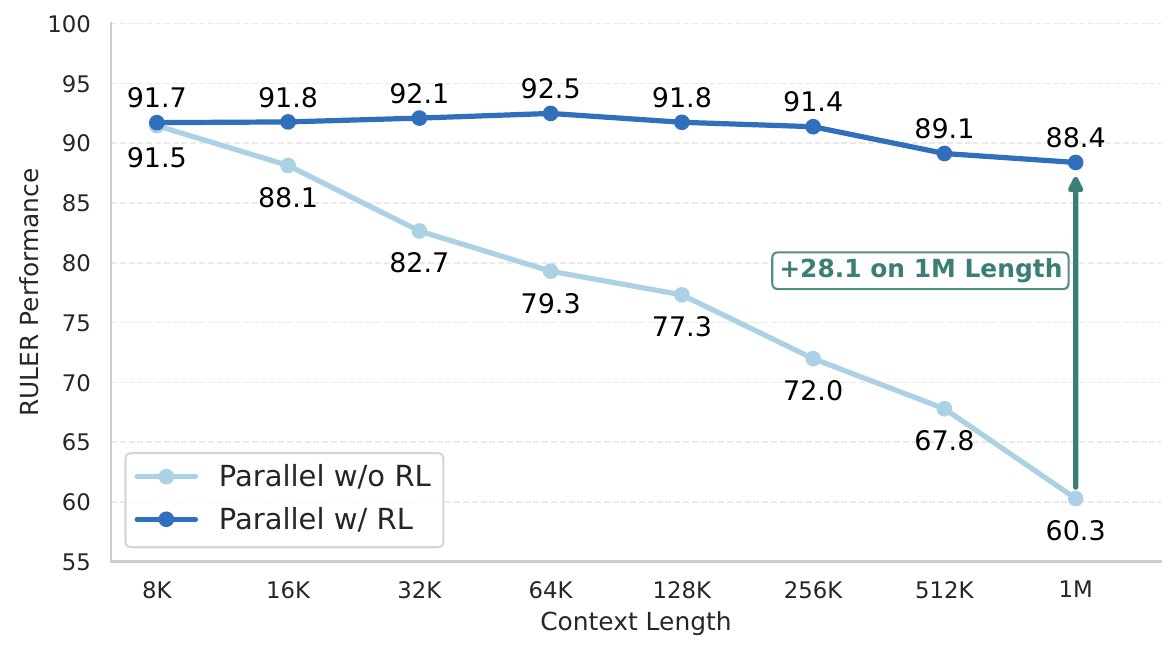}
\caption{Effect of RL training on RULER OOD performance across context lengths. RL training brings larger gains as the context becomes longer.}
\label{fig:ablation_rl_qwen25}
\end{figure}

\paragraph{Effect of RL Training.}
Figure~\ref{fig:ablation_rl_qwen25} evaluates whether RL training improves the proposed workflow on RULER OOD tasks. RL training yields only a minor gain at 8K tokens, but the improvement becomes increasingly clear as the context length grows. This trend suggests that even with simple synthetic data based on HQA, RL training helps the model generalize to more diverse long-context tasks and use the workflow more effectively.

\begin{figure}[t]
\centering
\includegraphics[width=\columnwidth]{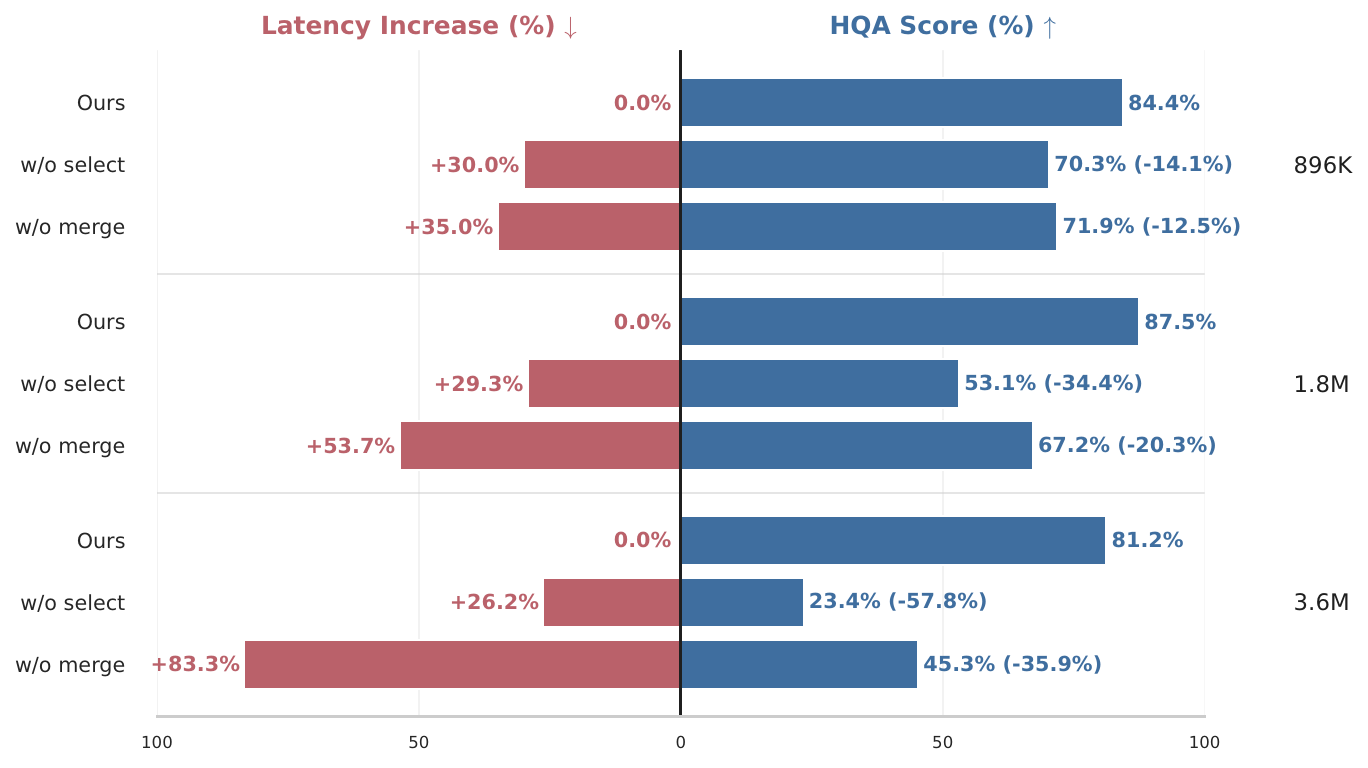}
\caption{Component ablation on HQA. HQA Score reports the absolute score, whereas Latency Increase reports the relative percentage increase over the full configuration. Removing either the select component or the merge component lowers HQA accuracy and increases latency, with the degradation becoming more severe at ultra-long context lengths.}
\label{fig:ablation_components_qwen25}
\end{figure}

\paragraph{Select and Merge Components.}
Figure~\ref{fig:ablation_components_qwen25} studies the select component and the merge component on HQA at ultra-long context lengths (896K, 1.8M, and 3.6M). In the variant without select, the model is not required to output a \texttt{\textless check\textgreater} signal, so all chunk-level observations are directly passed to the merge step. In the variant without merge, selected observations are simply concatenated across chunks rather than consolidated into a compact global memory.
Removing either component weakens context management: without selection, irrelevant, redundant, or noisy observations accumulate, whereas without merging, useful evidence remains fragmented and the memory expands rapidly.
As the input grows, the resulting context pollution or fragmentation leads to lower accuracy and longer inference time.

\begin{figure}[t]
\centering
\includegraphics[width=\columnwidth]{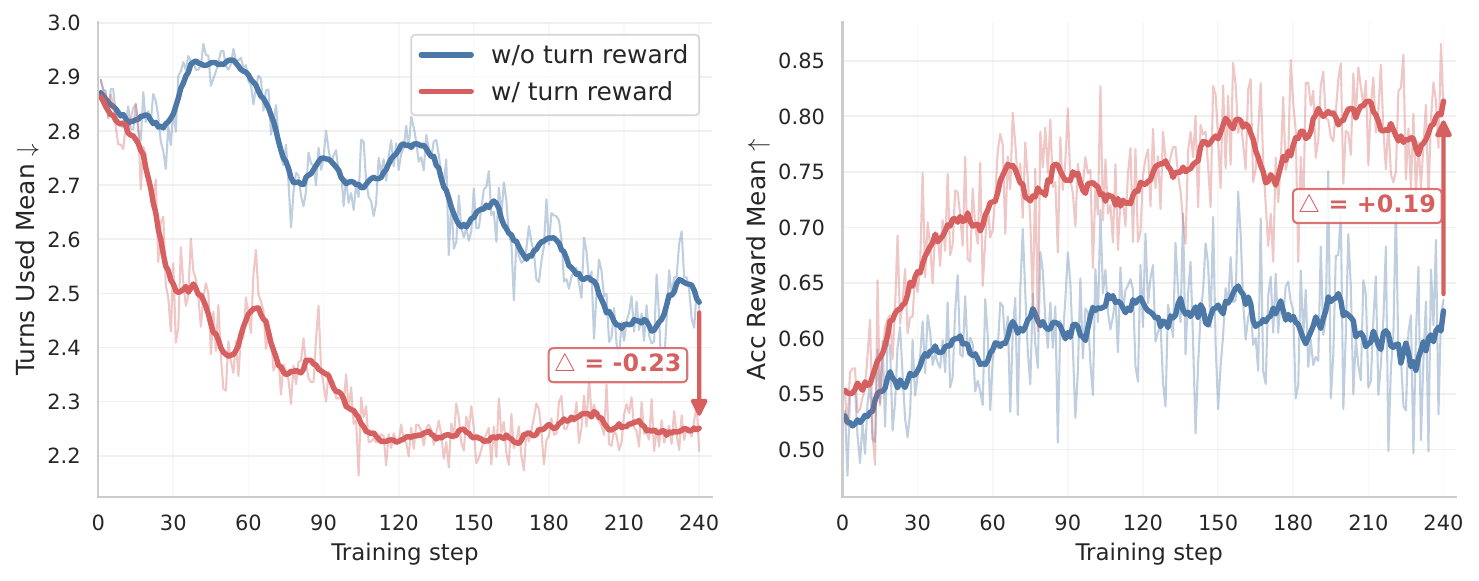}
\caption{Effect of the turn-efficiency reward on Qwen2.5-7B RL training. The turn reward leads to a faster, larger, and more stable reduction in the number of turns, while also improving the accuracy reward.}
\label{fig:ablation_turn_num}
\end{figure}

\paragraph{Turn-Efficiency Reward.}
Figure~\ref{fig:ablation_turn_num} compares training with and without the turn-efficiency reward; the reward-enabled variant uses $\lambda_{\mathrm{turn}}=0.2$.
The left panel shows that adding the turn reward produces a substantially larger reduction in the number of read-select-merge turns. Although the model trained without the turn reward also learns to use fewer turns as training proceeds, its curve is more unstable and the reduction is substantially smaller. By contrast, the turn-reward variant quickly suppresses redundant turns and converges to a lower and more stable turn count.
The right panel shows a higher late-stage accuracy reward, indicating that the efficiency gain preserves correctness. Before workflow-level training, extra turns can be unnecessary or harmful because repeated updates may introduce noise, overwrite useful evidence, or over-refine sufficient memory.

Further analyses in the Technical Supplement include more implementation details, additional ablation studies, actual turn counts, detailed comparisons of inference FLOPs and end-to-end latency, and comparative case studies.

\section{Conclusion}
We introduced PI-Mem, a parallel-iterative memory mechanism for long-context reasoning that gathers evidence from all chunks in parallel and iteratively refines a shared memory. Instead of updating memory recurrently after every chunk, PI-Mem reads chunks in parallel conditioned on a shared global memory, selects useful observations, and merges them into a compact memory across turns. We further post-trained the workflow with RL so the model can better follow the read-select-merge process and avoid unnecessary turns. Experiments with Qwen3.5-35B-A3B and Qwen2.5-7B demonstrate that PI-Mem improves long-context QA and retrieval performance while reducing inference latency compared with recurrent-memory and direct-inference baselines. Together, memory workflow design and end-to-end post-training offer a promising path to ultra-long-context reasoning without relying solely on larger native context windows.

\section*{Acknowledgments}
This work was supported by Shanghai Artificial Intelligence Laboratory and supported by the National Natural Science Foundation of China (Grant No. 6250076080) and supported by the China Postdoctoral Science Foundation under Grant Number
2025M771537.

\bibliography{aaai2027}



\onecolumn
\begingroup
\setcounter{secnumdepth}{2}
\setcounter{section}{0}
\setcounter{subsection}{0}
\setcounter{table}{0}
\setcounter{figure}{0}

\makeatletter
\def\section{\@startsection {section}{1}{\z@}{-2.0ex plus -0.5ex minus -.2ex}{12pt plus 2pt minus 1pt}{\Large\bf\centering}}
\def\subsection{\@startsection{subsection}{2}{\z@}{-2.0ex plus -0.5ex minus -.2ex}{3pt plus 2pt minus 1pt}{\large\bf\raggedright}}
\def\subsubsection{\@startsection{subparagraph}{3}{\z@}{-6pt plus -2pt minus -1pt}{-1em}{\normalsize\bf}}
\renewcommand\paragraph{\@startsection{paragraph}{4}{\z@}{-6pt plus -2pt minus -1pt}{-1em}{\normalsize\bf}}
\makeatother

\definecolor{pimemNoRL}{RGB}{250,243,250}
\definecolor{pimemRL}{RGB}{236,247,255}
\definecolor{error}{HTML}{d37a60}
\definecolor{errorlight}{HTML}{fdf8f7}
\definecolor{correct}{HTML}{518f5f}
\definecolor{correctlight}{HTML}{f5f9f6}
\definecolor{template}{HTML}{5c7ca5}
\definecolor{templatelight}{HTML}{f6f8fa}
\newcommand{\emphasizeerror}[1]{{\color{error}\textbf{#1}}}
\newcommand{\emphasizecorrect}[1]{{\color{correct}\textbf{#1}}}
\newcommand{\emphasizetemplate}[1]{{\color{template}\textbf{#1}}}

\vspace*{-0.21in}
\begin{center}
{\LARGE\bfseries Technical Supplement\par}
\end{center}
\vspace{0.5in}

\section{Training Details}

\begin{table}[h]
\centering
{
\begin{tabular}{l|cc}
\toprule
\textbf{Hyperparameter} & \textbf{Qwen3.5} & \textbf{Qwen2.5} \\
\midrule
Documents per sample & 1,000 & 200 \\
Context length & 140K & 28K \\
Chunk size & 15,000 & 5,000 \\
Maximum output length & 4,096 & 1,024 \\
Maximum turns ($K$) & \multicolumn{2}{c}{3} \\
Train batch size & \multicolumn{2}{c}{128} \\
Mini batch size & \multicolumn{2}{c}{8} \\
Adam optimizer $\beta_1,\beta_2$ & \multicolumn{2}{c}{$(0.9,\ 0.999)$} \\
Adam optimizer $\epsilon$ & \multicolumn{2}{c}{$1 \times 10^{-8}$} \\
Weight decay & \multicolumn{2}{c}{0.01} \\
Gradient clipping & \multicolumn{2}{c}{1.0} \\
Warmup steps & \multicolumn{2}{c}{20} \\
Learning rate & \multicolumn{2}{c}{$1 \times 10^{-6}$} \\
Learning-rate schedule & \multicolumn{2}{c}{Constant after linear warmup} \\
KL loss coefficient ($\beta$) & \multicolumn{2}{c}{$1 \times 10^{-3}$} \\
Clip ratio ($\varepsilon$) & \multicolumn{2}{c}{0.2} \\
MoE router auxiliary loss coef. & $1 \times 10^{-3}$ & N/A \\
GRPO group size ($G$) & 8 & 16 \\
Training rollout steps & 80 & 240 \\
Training samples & 10,240 & 30,720 \\
Temperature & \multicolumn{2}{c}{1.0} \\
Top-$p$ & \multicolumn{2}{c}{1.0} \\
Thinking mode & Disabled & N/A \\
GPU world size (H200) & 32 & 16 \\
Training Time (hours) & 75 & 96 \\
\bottomrule
\end{tabular}
}
\caption{Training hyperparameters for PI-Mem with Qwen3.5-35B-A3B and Qwen2.5-7B-Instruct.}
\label{supp:tab:training_hyperparameters}
\end{table}

Following MemAgent's released data-construction procedure, we synthesize HQA
training data by embedding the gold HotpotQA paragraphs into distractor
articles sampled from the same dataset. For Qwen2.5-7B, we use the
released MemAgent training data, in which each sample contains 200 articles
(approximately 28K tokens). Since Qwen3.5-35B-A3B has a longer native context
window, we additionally synthesize longer samples with the same construction
pipeline, using 1,000 articles (approximately 140K tokens) per sample. We
accordingly increase the chunk size from 5K to 15K tokens and the maximum output
length from 1,024 to 4,096 tokens.
Thinking mode is disabled for Qwen3.5 to improve rollout efficiency and reduce the generated output length of each model call.

Most RL hyperparameters and algorithmic choices follow the original MemAgent setup, including Dr.~GRPO-style advantage normalization (without division by the group standard deviation) and DAPO-style loss aggregation. We retain MemAgent's GRPO group size of 16 for Qwen2.5-7B, but reduce it to 8 for Qwen3.5-35B-A3B because
of the substantially higher training cost. For the Qwen3.5 comparison, we train
PI-Mem and MemAgent on the same longer dataset with identical hyperparameters.
For Qwen2.5-7B, we train only PI-Mem on the released MemAgent data and evaluate
the officially released MemAgent checkpoint.

\FloatBarrier

\section{Evaluation Details}

\subsection{Evaluation Setup}

\begin{table}[h]
\centering
{
\begin{tabular}{l|cc}
\toprule
\textbf{Configuration} & \textbf{Qwen3.5} & \textbf{Qwen2.5} \\
\midrule
Chunk size & 15,000 & 5,000 \\
Maximum output length & 4,096 & 1,024 \\
Maximum turns ($K$) & \multicolumn{2}{c}{3} \\
GPU world size (H200) & \multicolumn{2}{c}{8} \\
Tensor parallel size & \multicolumn{2}{c}{2} \\
Samples per subtask (RULER) & \multicolumn{2}{c}{64} \\
Thinking mode & Disabled & N/A \\
Temperature & \multicolumn{2}{c}{0.7} \\
Top-$p$ & \multicolumn{2}{c}{0.95} \\
\bottomrule
\end{tabular}
}
\caption{Evaluation configurations for PI-Mem with Qwen3.5-35B-A3B and Qwen2.5-7B-Instruct.}
\label{supp:tab:evaluation_configurations}
\end{table}

Following MemAgent, single-answer tasks use normalized answer matching (Sub-EM), whereas multi-answer tasks are scored by the fraction of target values appearing in the prediction; LongBench v2 follows its official evaluation protocol.
The evaluation temperature and top-$p$ also follow MemAgent's official code.
Each model reported in this work is trained in a single run, and each evaluation sample is evaluated once per method.

\paragraph{YaRN.}
We use a scaling factor of 4.0 for positional-encoding extrapolation.

\paragraph{RAG.}
For each evaluation sample, we construct an Okapi BM25 index over its decoded text chunks; the index is therefore sample-specific rather than a shared external knowledge base. We use the original question as the retrieval query and select the top-6 chunks in descending order of their BM25 scores.

\FloatBarrier
\subsection{Detailed Evaluation Results}

\begin{table}[!htbp]
\centering
{
\begin{tabular}{l|cccccccccc|c}
\toprule
Method & 7K & 14K & 28K & 56K & 112K & 224K & 448K & 896K & 1.8M & 3.6M & Avg. \\
\midrule
\multicolumn{12}{c}{\textit{\textbf{Qwen3.5-35B-A3B}}} \\
\midrule
Vanilla & \textbf{87.50} & 79.69 & 82.81 & 81.25 & 78.12 & 73.44 & 62.50 & 37.50 & 34.38 & 29.69 & 64.69 \\
YaRN & 82.81 & 81.25 & 79.69 & \textbf{82.81} & 78.12 & 68.75 & 59.38 & 56.25 & 29.69 & 14.06 & 63.28 \\
RAG & 84.38 & \textbf{82.81} & 71.88 & 67.19 & 46.88 & 54.69 & 51.56 & 53.12 & 46.88 & 43.75 & 60.31 \\
MemAgent (w/o RL) & 78.12 & 78.12 & 70.31 & 71.88 & 62.50 & 60.94 & 62.50 & 68.75 & 51.56 & 39.06 & 64.37 \\
MemAgent & 84.38 & 81.25 & 82.81 & 81.25 & 79.69 & 68.75 & 73.44 & 71.88 & 71.88 & 70.31 & 76.56 \\
\rowcolor{pimemNoRL}PI-Mem (w/o RL) & 78.12 & 75.00 & 73.44 & 68.75 & 67.19 & 67.19 & 71.88 & 64.06 & 65.62 & 67.19 & 69.84 \\
\rowcolor{pimemRL}\textbf{PI-Mem} & \textbf{87.50} & \textbf{82.81} & \textbf{84.38} & \textbf{82.81} & \textbf{84.38} & \textbf{79.69} & \textbf{82.81} & \textbf{76.56} & \textbf{75.00} & \textbf{76.56} & \textbf{81.25} \\
\midrule
\rowcolor{pimemNoRL}PI-Mem (w/o RL) Turns & 2.133 & 2.125 & 2.148 & 2.227 & 2.180 & 2.227 & 2.258 & 2.258 & 2.188 & 2.344 & 2.209 \\
\rowcolor{pimemRL}\textbf{PI-Mem} Turns & 2.016 & 2.016 & 2.047 & 2.047 & 2.062 & 2.078 & 2.078 & 2.094 & 2.094 & 2.156 & 2.069 \\
\midrule
\multicolumn{12}{c}{\textit{\textbf{Qwen2.5-7B}}} \\
\midrule
Vanilla & 60.94 & 54.69 & 50.00 & 23.44 & 0.00 & 0.00 & 0.00 & 0.00 & 0.00 & 0.00 & 18.91 \\
YaRN & 56.25 & 62.50 & 62.50 & 45.31 & 34.38 & 0.00 & 0.00 & 0.00 & 0.00 & 0.00 & 26.09 \\
RAG & 57.81 & 46.88 & 46.88 & 45.31 & 39.06 & 42.19 & 29.69 & 43.75 & 26.56 & 32.81 & 41.09 \\
GRU-Mem & 81.25 & 79.69 & 77.34 & 76.56 & 74.22 & 73.44 & 71.88 & 76.56 & N/A & N/A & 76.37 \\
ReMemR1 & 82.3 & 82.8 & 81.1 & 78.9 & 82.0 & 79.7 & 80.0 & 80.8 & N/A & N/A & 80.95 \\
MemAgent (w/o RL) & 62.50 & 59.38 & 51.56 & 50.00 & 43.75 & 40.62 & 35.94 & 37.50 & 31.25 & 32.81 & 44.53 \\
MemAgent & 81.25 & 81.25 & 75.00 & \textbf{82.81} & 76.56 & 75.00 & 76.56 & 75.00 & 78.12 & 73.44 & 77.50 \\
\rowcolor{pimemNoRL}PI-Mem (w/o RL) & 57.03 & 51.56 & 47.66 & 44.53 & 41.41 & 46.46 & 34.38 & 35.94 & 35.94 & 28.12 & 42.30 \\
\rowcolor{pimemRL}\textbf{PI-Mem} & \textbf{82.81} & \textbf{82.81} & \textbf{81.25} & \textbf{82.81} & \textbf{89.06} & \textbf{85.94} & \textbf{82.81} & \textbf{84.38} & \textbf{87.50} & \textbf{81.25} & \textbf{84.06} \\
\midrule
\rowcolor{pimemNoRL}PI-Mem (w/o RL) Turns & 2.883 & 2.875 & 2.859 & 2.875 & 2.859 & 2.921 & 2.930 & 2.938 & 2.891 & 2.906 & 2.894 \\
\rowcolor{pimemRL}\textbf{PI-Mem} Turns & 2.312 & 2.203 & 2.156 & 2.219 & 2.172 & 2.188 & 2.234 & 2.234 & 2.234 & 2.328 & 2.228 \\
\bottomrule
\end{tabular}
}
\caption{Length-grouped RULER HQA results. Scores and mean turn counts are reported. The averages for GRU-Mem and ReMemR1 use the eight reported lengths.}
\label{supp:tab:ruler_hqa_length_results}
\end{table}

\begin{figure}[!htbp]
\centering
\includegraphics[width=0.94\textwidth]{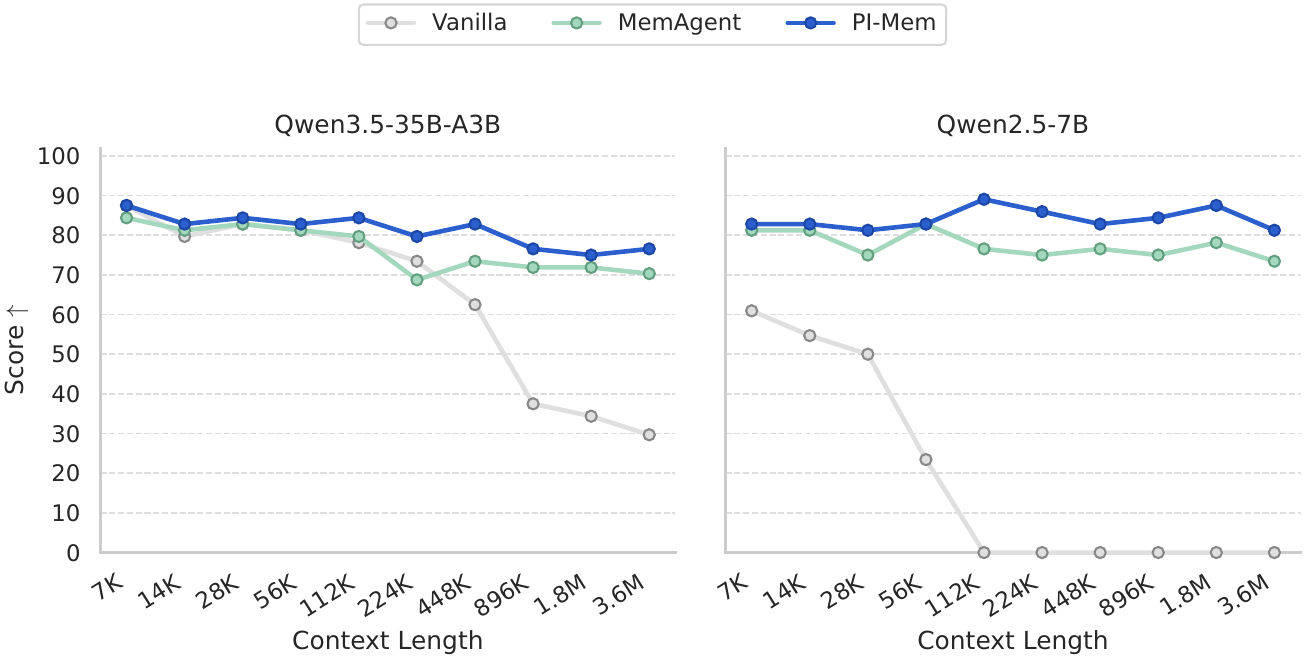}
\caption{HQA performance across context lengths for Qwen3.5-35B-A3B (left) and Qwen2.5-7B (right).}
\label{supp:fig:hqa_performance_by_length}
\end{figure}

\begin{table}[!htbp]
\centering
{
\setlength{\tabcolsep}{9pt}
\begin{tabular}{l|cccccccc|c}
\toprule
Method & 8K & 16K & 32K & 64K & 128K & 256K & 512K & 1M & Avg. \\
\midrule
\multicolumn{10}{c}{\textit{\textbf{Qwen3.5-35B-A3B}}} \\
\midrule
Vanilla & 98.67 & 98.39 & 97.64 & 97.96 & 97.40 & 96.08 & 95.81 & 75.56 & 94.69 \\
YaRN & 97.92 & 96.59 & 96.74 & 97.02 & 97.46 & 97.22 & 96.19 & 86.58 & 95.72 \\
RAG & 98.31 & 93.07 & 94.32 & 95.18 & 93.25 & 87.02 & 77.90 & 75.24 & 89.29 \\
MemAgent (w/o RL) & 95.50 & 88.91 & 77.38 & 63.93 & 56.76 & 50.25 & 50.03 & 39.32 & 65.26 \\
MemAgent & 94.55 & 92.92 & 95.18 & 94.05 & 92.46 & 90.84 & 86.42 & 83.86 & 91.28 \\
\rowcolor{pimemNoRL}PI-Mem (w/o RL) & 97.03 & 95.66 & 96.87 & 96.86 & 96.43 & 96.27 & 95.71 & 95.15 & 96.25 \\
\rowcolor{pimemRL}\textbf{PI-Mem} & \textbf{98.92} & \textbf{98.65} & \textbf{98.06} & \textbf{98.42} & \textbf{97.69} & \textbf{98.38} & \textbf{97.30} & \textbf{96.88} & \textbf{98.04} \\
\midrule
\rowcolor{pimemNoRL}PI-Mem (w/o RL) Turns & 2.043 & 2.071 & 2.163 & 2.204 & 2.212 & 2.225 & 2.274 & 2.294 & 2.186 \\
\rowcolor{pimemRL}\textbf{PI-Mem} Turns & 2.055 & 2.044 & 2.041 & 2.021 & 2.024 & 2.037 & 2.065 & 2.068 & 2.044 \\
\midrule
\multicolumn{10}{c}{\textit{\textbf{Qwen2.5-7B}}} \\
\midrule
Vanilla & 86.57 & 87.17 & 84.48 & 60.65 & 22.72 & 0.00 & 0.00 & 0.00 & 42.70 \\
YaRN & 84.24 & 83.03 & 78.82 & 72.48 & 57.13 & 0.00 & 0.00 & 0.00 & 46.96 \\
RAG & 80.58 & 79.30 & 77.57 & 76.49 & 74.07 & 68.05 & 63.45 & 58.70 & 72.28 \\
MemAgent (w/o RL) & \textbf{94.22} & \textbf{91.92} & 88.35 & 85.96 & 81.16 & 78.19 & 72.22 & 62.41 & 81.80 \\
MemAgent & 91.66 & 90.33 & 89.31 & 87.73 & 83.73 & 80.30 & 79.51 & 74.83 & 84.68 \\
\rowcolor{pimemNoRL}PI-Mem (w/o RL) & 91.47 & 88.14 & 82.66 & 79.30 & 77.32 & 71.99 & 67.80 & 60.29 & 77.37 \\
\rowcolor{pimemRL}\textbf{PI-Mem} & 91.72 & 91.78 & \textbf{92.10} & \textbf{92.49} & \textbf{91.75} & \textbf{91.38} & \textbf{89.14} & \textbf{88.39} & \textbf{91.09} \\
\midrule
\rowcolor{pimemNoRL}PI-Mem (w/o RL) Turns & 2.825 & 2.852 & 2.864 & 2.905 & 2.891 & 2.896 & 2.920 & 2.920 & 2.884 \\
\rowcolor{pimemRL}\textbf{PI-Mem} Turns & 2.038 & 2.057 & 2.057 & 2.038 & 2.044 & 2.065 & 2.098 & 2.101 & 2.062 \\
\bottomrule
\end{tabular}
}
\caption{Length-grouped RULER out-of-distribution results. Scores are reported across context lengths from 8K to 1M tokens; mean turn counts are additionally averaged over the 11 out-of-distribution tasks at each length.}
\label{supp:tab:ruler_ood_length_results}
\end{table}

\begin{figure}[!htbp]
\centering
\includegraphics[width=0.94\textwidth]{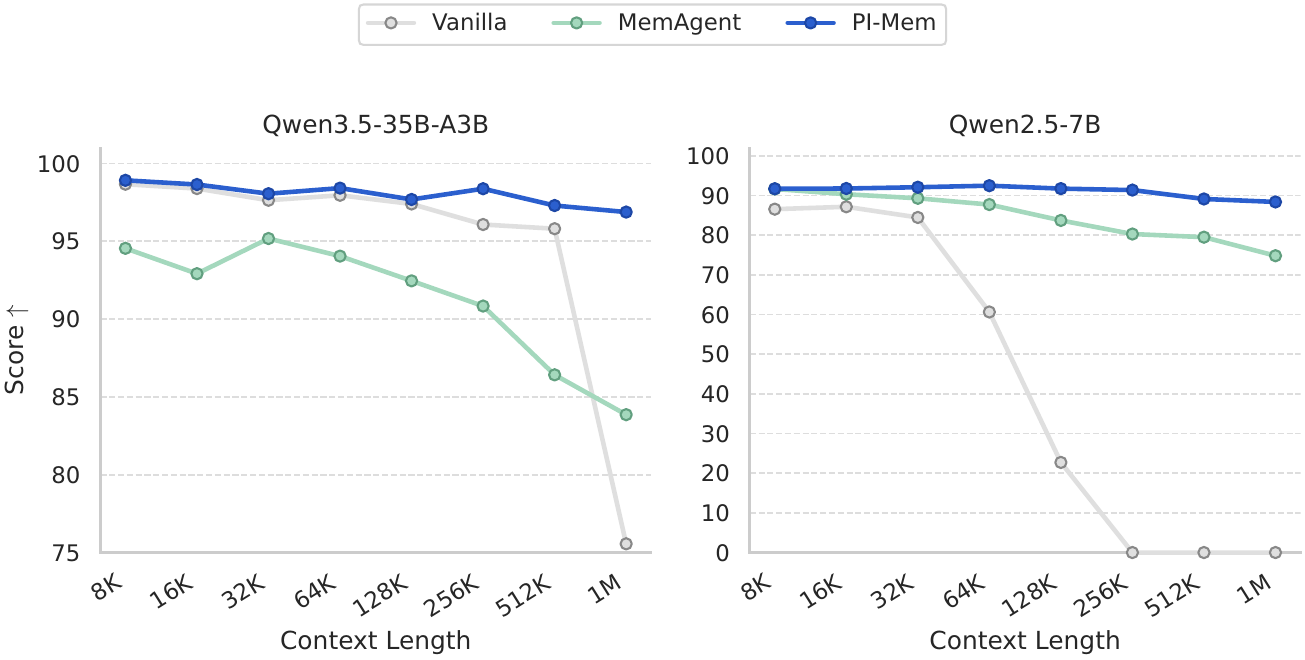}
\caption{RULER OOD performance across context lengths for Qwen3.5-35B-A3B (left) and Qwen2.5-7B (right).}
\label{supp:fig:ood_performance_by_length}
\end{figure}

\begin{table}[!htbp]
\centering
{
\setlength{\tabcolsep}{8pt}
\begin{tabular}{l|c|cc|ccc}
\toprule
\multirow{2}{*}{\textbf{Method}} &
\multirow{2}{*}{\textbf{All}} &
\multicolumn{2}{c|}{\textbf{Difficulty}} &
\multicolumn{3}{c}{\textbf{Length}} \\
& & Easy & Hard & Short & Medium & Long \\
\midrule
Vanilla & 50.7 & 55.7 & 47.6 & 56.1 & 47.9 & 47.2 \\
YaRN & 51.5 & 52.1 & \textbf{51.1} & \textbf{57.8} & 48.4 & 47.2 \\
RAG & 48.9 & 53.6 & 46.0 & 51.7 & 44.7 & 52.8 \\
MemAgent (w/o RL) & 49.3 & 55.2 & 45.7 & 53.9 & 46.0 & 48.1 \\
MemAgent& 52.3 & 59.9 & 47.6 & 55.6 & 48.8 & 53.7 \\
\rowcolor{pimemNoRL}PI-Mem (w/o RL) & 51.7 & 63.5 & 44.4 & 51.7 & 48.4 & 58.3 \\
\rowcolor{pimemRL}\textbf{PI-Mem} & \textbf{54.1} & \textbf{66.1} & 46.6 & 53.3 & \textbf{49.8} & \textbf{63.9} \\
\bottomrule
\end{tabular}
}
\caption{LongBench v2 results using Qwen3.5-35B-A3B. Following the grouping defined by the benchmark, we report results by difficulty (Easy and Hard) and context length (Short, Medium, and Long).}
\label{supp:tab:longbenchv2_results}
\end{table}

\FloatBarrier
\subsection{Computational Cost and End-to-End Latency}

Table~\ref{supp:tab:flops_latency} compares Vanilla, MemAgent, and PI-Mem on the 64-sample HQA evaluation sets from 112K to 3.6M tokens. We report analytically estimated algorithmic FLOPs, counting one multiply--accumulate operation (MAC) as two FLOPs; PFLOPs denotes the aggregate operation count in units of $10^{15}$ FLOPs rather than throughput. For PI-Mem, the estimate includes all chunk reads, refinement turns, memory updates, and final-answer generation, thereby covering the complete inference workload. Wall-clock latency is measured end-to-end for the same 64 samples under the evaluation configuration in Table~\ref{supp:tab:evaluation_configurations}. The speedup ratio is computed as the Vanilla latency divided by the latency of each method.

For a workflow-level asymptotic comparison, let $N$ denote the number of chunks, $C$ the chunk size, $M$ the fixed memory length, and $K$ the maximum number of refinement turns; the total input length is therefore $NC$. Table~\ref{supp:tab:asymptotic_cost} reports the dominant chunk-reading attention cost under standard quadratic self-attention. The comparison omits merge and final-answer costs to isolate the effect of workflow structure; these components are included in the complete PFLOPs estimates in Table~\ref{supp:tab:flops_latency}.

\begin{table}[!htbp]
\centering
{
\begin{tabular}{lcc}
\toprule
\textbf{Method} & \textbf{Attention FLOPs} & \textbf{Serial Depth} \\
\midrule
Vanilla & $\mathcal{O}((NC)^2)$ & $1$ \\
MemAgent & $\mathcal{O}(N(C+M)^2)$ & $\mathcal{O}(N)$ \\
\rowcolor{controlled}PI-Mem & $\mathcal{O}(KN(C+M)^2)$ & $\mathcal{O}(K)$ \\
\bottomrule
\end{tabular}
}
\caption{Asymptotic chunk-reading attention cost and serial depth under quadratic self-attention. Vanilla uses one ultra-long call, whereas PI-Mem processes all $N$ chunk reads in parallel within each turn.}
\label{supp:tab:asymptotic_cost}
\end{table}

The asymptotic analysis exposes a compute--latency trade-off for PI-Mem: relative to recurrent memory, it performs up to $K$ times more chunk-reading work but reduces the sequential critical path from $\mathcal{O}(N)$ to $\mathcal{O}(K)$. Because $K$ is bounded and typically much smaller than $N$ for ultra-long inputs, PI-Mem converts additional computation into parallelizable work. The empirical results indicate that this trade-off is most favorable in the ultra-long-context regime, where increased parallelism outweighs the additional computation.

\begin{table}[!htbp]
\centering
{
\begin{tabular}{c|l|rrr|rrr}
\toprule
\multirow{2}{*}{\textbf{Length}} &
\multirow{2}{*}{\textbf{Method}} &
\multicolumn{3}{c|}{\textit{\textbf{Qwen3.5-35B-A3B}}} &
\multicolumn{3}{c}{\textit{\textbf{Qwen2.5-7B}}} \\
& & \textbf{PFLOPs} $\downarrow$ & \textbf{Latency (s)} $\downarrow$ & \textbf{Speedup} $\uparrow$ &
\textbf{PFLOPs} $\downarrow$ & \textbf{Latency (s)} $\downarrow$ & \textbf{Speedup} $\uparrow$ \\
\midrule
\multirow{3}{*}{112K} & Vanilla & 103.98 & 62.94 & 1.00$\times$ & 259.70 & 80.41 & 1.00$\times$ \\
 & MemAgent & 57.53 & 258.89 & 0.24$\times$ & 129.70 & 124.58 & 0.65$\times$ \\
\rowcolor{pimemRL}\cellcolor{white} & PI-Mem & 101.01 & 151.72 & 0.41$\times$ & 226.46 & 80.12 & 1.00$\times$ \\
\addlinespace[2pt]
\multirow{3}{*}{224K} & Vanilla & 344.67 & 133.17 & 1.00$\times$ & 854.33 & 319.69 & 1.00$\times$ \\
 & MemAgent & 121.94 & 627.33 & 0.21$\times$ & 267.99 & 300.10 & 1.07$\times$ \\
\rowcolor{pimemRL}\cellcolor{white} & PI-Mem & 201.07 & 164.14 & 0.81$\times$ & 454.48 & 154.72 & 2.07$\times$ \\
\addlinespace[2pt]
\multirow{3}{*}{448K} & Vanilla & 1,232.93 & 416.74 & 1.00$\times$ & 3,040.02 & 849.50 & 1.00$\times$ \\
 & MemAgent & 249.47 & 1,243.47 & 0.34$\times$ & 543.88 & 629.07 & 1.35$\times$ \\
\rowcolor{pimemRL}\cellcolor{white} & PI-Mem & 401.34 & 317.06 & 1.31$\times$ & 909.09 & 309.65 & 2.74$\times$ \\
\addlinespace[2pt]
\multirow{3}{*}{896K} & Vanilla & 4,638.38 & 1,341.35 & 1.00$\times$ & 11,404.64 & 3,176.89 & 1.00$\times$ \\
 & MemAgent & 506.60 & 2,793.83 & 0.48$\times$ & 1,099.70 & 1,308.64 & 2.43$\times$ \\
\rowcolor{pimemRL}\cellcolor{white} & PI-Mem & 806.86 & 500.39 & 2.68$\times$ & 1,832.27 & 625.76 & 5.08$\times$ \\
\addlinespace[2pt]
\multirow{3}{*}{1.8M} & Vanilla & 18,041.40 & 13,237.88 & 1.00$\times$ & 44,280.96 & 10,112.76 & 1.00$\times$ \\
 & MemAgent & 1,020.13 & 5,848.31 & 2.26$\times$ & 2,215.06 & 2,634.93 & 3.84$\times$ \\
\rowcolor{pimemRL}\cellcolor{white} & PI-Mem & 1,615.29 & 984.08 & 13.45$\times$ & 3,671.00 & 1,285.31 & 7.87$\times$ \\
\addlinespace[2pt]
\multirow{3}{*}{3.6M} & Vanilla & 71,573.35 & 33,365.14 & 1.00$\times$ & 175,515.76 & 38,653.00 & 1.00$\times$ \\
 & MemAgent & 2,054.65 & 12,060.71 & 2.77$\times$ & 4,462.91 & 5,535.50 & 6.98$\times$ \\
\rowcolor{pimemRL}\cellcolor{white} & PI-Mem & 3,365.35 & 1,974.65 & 16.90$\times$ & 7,633.65 & 2,682.48 & 14.41$\times$ \\
\bottomrule
\end{tabular}
}
\caption{Aggregate algorithmic FLOPs and end-to-end wall-clock latency for 64 HQA samples. PFLOPs denotes $10^{15}$ floating-point operations. FLOPs are analytically estimated, while latency is measured. Speedup is relative to Vanilla.}
\label{supp:tab:flops_latency}
\end{table}

\clearpage
\section{Additional Ablations}

\subsection{Maximum Turn Budget}

\begin{table}[!htbp]
\centering
{
\begin{tabular}{c l|cccccccccc}
\toprule
$K$ & Metric & 7K & 14K & 28K & 56K & 112K & 224K & 448K & 896K & 1.8M & 3.6M \\
\midrule
\multirow{2}{*}{1}
 & HQA score & 73.44 & 71.88 & 75.00 & 75.00 & 78.12 & 79.69 & 67.19 & 76.56 & 67.19 & 70.31 \\
 & Mean turns used & 1.000 & 1.000 & 1.000 & 1.000 & 1.000 & 1.000 & 1.000 & 1.000 & 1.000 & 1.000 \\
\midrule
\multirow{2}{*}{3}
 & HQA score & 82.81 & 82.81 & \textbf{81.25} & \textbf{82.81} & 89.06 & \textbf{85.94} & 82.81 & \textbf{84.38} & \textbf{87.50} & \textbf{81.25} \\
 & Mean turns used & 2.312 & 2.203 & 2.156 & 2.219 & 2.172 & 2.188 & 2.234 & 2.234 & 2.234 & 2.328 \\
\midrule
\multirow{2}{*}{5}
 & HQA score & \textbf{84.38} & \textbf{84.38} & 79.69 & 79.69 & \textbf{90.62} & \textbf{85.94} & \textbf{84.38} & \textbf{84.38} & 85.94 & \textbf{81.25} \\
 & Mean turns used & 2.375 & 2.344 & 2.109 & 2.281 & 2.312 & 2.234 & 2.359 & 2.234 & 2.375 & 2.484 \\
\bottomrule
\end{tabular}
}
\caption{Effect of the maximum turn budget $K$ on Qwen2.5-7B HQA performance and the mean number of turns used. Bold values denote the best HQA score at each context length; ties are all highlighted.}
\label{supp:tab:max_turn_ablation}
\end{table}

Table~\ref{supp:tab:max_turn_ablation} shows that limiting PI-Mem to one turn consistently reduces HQA accuracy. With $K=1$, all chunk reads are conditioned only on the initial empty memory, so evidence discovered in one chunk cannot guide subsequent reads of other chunks; iterative refinement is therefore important for cross-chunk information exchange. Increasing the budget from $K=3$ to $K=5$, however, does not yield consistent further gains because most samples converge early: under $K=5$, 75.94\% of samples exit after two turns, whereas only 4.84\% use four or five turns. PI-Mem thus benefits from iteration while invoking additional turns only for the small subset of samples that require further refinement.

\subsection{Chunk and Memory Sizes}

We evaluate training-free PI-Mem with Qwen3.5-35B-A3B on HQA while varying
one size parameter at a time. The default configuration uses 15K-token chunks
and a 4K-token memory. For the chunk-size ablation, the memory size remains
4K; for the memory-size ablation, the chunk size remains 15K.

\begin{table}[!htbp]
\centering
{
\begin{tabular}{cc|cccccccccc}
\toprule
Chunk & Memory & 7K & 14K & 28K & 56K & 112K & 224K & 448K & 896K & 1.8M & 3.6M \\
\midrule
5K  & 4K & 71.88 & 70.31 & 65.62 & 70.31 & 67.19 & 70.31 & 62.50 & 67.19 & 65.62 & 65.62 \\
15K & 2K & 78.12 & 76.56 & 67.19 & 68.75 & 59.38 & 65.62 & 64.06 & 70.31 & 65.62 & 59.38 \\
\rowcolor{controlled}15K & 4K & 78.12 & 75.00 & 73.44 & 68.75 & 67.19 & 67.19 & 71.88 & 64.06 & 65.62 & 67.19 \\
15K & 8K & 78.12 & 76.56 & 70.31 & 70.31 & 65.62 & 64.06 & 70.31 & 68.75 & 60.94 & 67.19 \\
25K & 4K & 78.12 & 73.44 & 67.19 & 75.00 & 70.31 & 67.19 & 64.06 & 71.88 & 62.50 & 67.19 \\
\bottomrule
\end{tabular}
}
\caption{Effect of chunk and memory sizes on training-free PI-Mem with Qwen3.5-35B-A3B on HQA. One parameter is varied at a time, and the shaded row denotes the default 15K/4K configuration.}
\label{supp:tab:chunk_memory_ablation}
\end{table}

Table~\ref{supp:tab:chunk_memory_ablation} shows no consistent degradation across
context lengths when the chunk size varies from 5K to 25K or the memory size
varies from 2K to 8K.
PI-Mem is therefore robust to these size choices, and
its performance is not materially affected within the tested ranges. We do
not conduct an additional size ablation for Qwen2.5-7B
because its chunk and memory sizes exactly follow the experimental setup of
MemAgent.

\clearpage
\section{Prompt Template}

\tcbset{
  prompttemplate/.style={
    colback=templatelight,
    colframe=template,
    coltitle=white,
    width=0.9\linewidth,
    center,
    arc=5mm,
    fontupper=\carlito,
    fonttitle=\carlito\bfseries
  }
}

\begin{tcolorbox}[prompttemplate,title={Prompt Template for \textsc{ReadCall}}]
You are presented with a problem, a section of an article, and a global memory summarizing previously gathered information. Please read the section carefully and determine whether the section contains new information relevant to answering the problem beyond what is already in the global memory. First, output your judgment in the format \textbf{\textless check\textgreater}yes\textbf{\textless/check\textgreater} if there is new information, or \textbf{\textless check\textgreater}no\textbf{\textless/check\textgreater} if there is none. Then, if there is new information, extract and list the key details.

\medskip
\textbf{\textless problem\textgreater}\par
\{prompt\}\par
\textbf{\textless/problem\textgreater}

\medskip
\textbf{\textless memory\textgreater}\par
\{memory\}\par
\textbf{\textless/memory\textgreater}

\medskip
\textbf{\textless section\textgreater}\par
\{chunk\}\par
\textbf{\textless/section\textgreater}

\medskip
Your response:
\end{tcolorbox}

\begin{tcolorbox}[prompttemplate,title={Prompt Template for \textsc{MergeCall}}]
You are presented with a problem and key information extracted from multiple sections of an article. Please consolidate all the information into a single comprehensive memory. Remove redundancies and organize the information clearly, retaining all details relevant to answering the problem.

\medskip
\textbf{\textless problem\textgreater}\par
\{prompt\}\par
\textbf{\textless/problem\textgreater}

\medskip
\textbf{\textless extracted\_information\textgreater}\par
\{memories\}\par
\textbf{\textless/extracted\_information\textgreater}

\medskip
Consolidated memory:
\end{tcolorbox}

\begin{tcolorbox}[prompttemplate,title={Prompt Template for \textsc{FinalCall}}]
You are presented with a problem and a previous memory. Please answer the problem based on the previous memory and put the answer in \textbackslash boxed\{\{\}\}.

\medskip
\textbf{\textless problem\textgreater}\par
\{prompt\}\par
\textbf{\textless/problem\textgreater}

\medskip
\textbf{\textless memory\textgreater}\par
\{memory\}\par
\textbf{\textless/memory\textgreater}

\medskip
Your answer:
\end{tcolorbox}

\clearpage
\section{Case Studies}

We present paired Qwen3.5-35B-A3B trajectories on the same questions. The
MV-NIAH example has 130,778 tokens, and the HQA example has 1,600 documents
and 226,114 tokens. We retain the terminology used in the main paper: one
PI-Mem \emph{turn} is a complete read--select--merge cycle. Inside the colored
boxes, all prose is transcribed from the corresponding JSONL fields; we add
only structural labels and color highlighting. Ellipses mark omitted portions
of long outputs.

\tcbset{
  casememagent/.style={
    colback=errorlight,
    colframe=error,
    coltitle=white,
    width=0.9\linewidth,
    center,
    arc=4mm,
    boxrule=0.8pt,
    fontupper=\carlito\footnotesize,
    fonttitle=\carlito\bfseries
  },
  casepimem/.style={
    colback=correctlight,
    colframe=correct,
    coltitle=white,
    width=0.9\linewidth,
    center,
    arc=4mm,
    boxrule=0.8pt,
    fontupper=\carlito\footnotesize,
    fonttitle=\carlito\bfseries
  }
}

\newcommand{\casecheckyes}{\emphasizecorrect{\textless check\textgreater yes\textless/check\textgreater}}
\newcommand{\casecheckno}{\textbf{\textless check\textgreater no\textless/check\textgreater}}

\subsection{Case 1: MemAgent on MV-NIAH}

\begin{tcolorbox}[casememagent,title={MemAgent on MV-NIAH}]
\textbf{Question:} Some special magic numbers are hidden within the following
text. Make sure to memorize it. What are all the special magic numbers for
\texttt{cloudy-singing} mentioned in the provided text?

\smallskip\hrule\smallskip
\textbf{Chunk 0:}
“One of the special magic numbers for cloudy-singing is:
\emphasizecorrect{2952447}. July 2010What hard liquor, cigarettes, heroin,
and crack have in common is that they're all more concentrated forms of less
addictive predecessors. Most if not all the things we describe as addictive
are. And the scary thing is, the process that created them is accelerating.”

\smallskip
\textbf{Memory 0:}
“Based on the provided text, here is the updated memory containing the special
magic number for ‘cloudy-singing’ and other relevant details from the article:

\textbf{Updated Memory:}
\begin{itemize}
    \item \textbf{Special Magic Number for Cloudy-Singing:}
    \emphasizecorrect{2952447}
    \item \textbf{Context of the Number:} The number appears in the text
    immediately following the phrase ‘One of the special magic numbers for
    cloudy-singing is:’ and is followed by the date ‘July 2010’.
    \item \textbf{Article Themes \& Key Concepts:}
    \begin{itemize}
        \item \textbf{Addictiveness \& Technological Progress:} The text
        discusses how technological progress accelerates the concentration of
        things we like (e.g., opium to heroin, food processing, social media),
        making the world more addictive. It notes that social customs
        (antibodies) evolve slowly compared to technological progress,
        creating a need for individuals to figure out what to avoid on their
        own.
        \item \ldots
    \end{itemize}
\end{itemize}
”

\begin{center}\ldots\end{center}

\textbf{Chunk 2:}
“One of the special magic numbers for cloudy-singing is:
\emphasizecorrect{8372717}.”

\smallskip
\textbf{Memory 2:}
“Based on the provided text, here is the updated memory containing the new
special magic number for ‘cloudy-singing’ and all other relevant details from
the article:

\textbf{Updated Memory:}
\begin{itemize}
    \item \textbf{Special Magic Number for Cloudy-Singing:}
    \emphasizecorrect{8372717}
    \begin{itemize}
        \item \emphasizeerror{\emph{Note:} The previous memory listed
        2952447.} The text explicitly states: ‘One of the special magic
        numbers for cloudy-singing is: 8372717.’ This new number appears in the
        section discussing how technology makes mass-produced goods (like cars
        and watches) convenient and affordable for the rich, contrasting them
        with expensive, custom-made items that are often inferior in
        performance.
    \end{itemize}
    \item \textbf{Context of the Number:} The number appears in the essay
    ‘Make Money as a Specialized Skill’ (April 2004), specifically within a
    paragraph comparing expensive custom cars to mass-produced cars, and
    expensive mechanical watches to accurate quartz watches.
\end{itemize}
\ldots”
\end{tcolorbox}

\begin{tcolorbox}[casememagent,title={MemAgent on MV-NIAH --- Continued}]
\textbf{Chunk 6:}
“One of the special magic numbers for cloudy-singing is:
\emphasizecorrect{8023581}.”

\smallskip
\textbf{Memory 6:}
“Based on the provided text, here is the updated memory containing the new
special magic number for ‘cloudy-singing’ and all other relevant details from
the article:

\textbf{Updated Memory:}
\begin{itemize}
    \item \textbf{Special Magic Number for Cloudy-Singing:}
    \emphasizecorrect{8023581}
    \begin{itemize}
        \item \emphasizeerror{\emph{Note:} The previous memory listed
        8372717 (and originally 2952447).} The text explicitly states: ‘One of
        the special magic numbers for cloudy-singing is: 8023581.’ This new
        number appears in the section discussing ‘Rounds’ and the structure of
        angel rounds versus series A rounds.
    \end{itemize}
    \item \textbf{Context of the Number:} The number appears in the essay
    ‘How to Start a Startup’ (April 2012), specifically within a paragraph
    comparing the equity dilution of traditional Series A rounds (30--40\%)
    with the lower dilution of angel rounds (often around 13\%).
\end{itemize}
\ldots

\textbf{Correction to Previous Memory:}
\begin{itemize}
    \item The previous memory listed \textbf{8372717} as the number for
    ‘cloudy-singing’ found in the ‘Make Money as a Specialized Skill’ essay.
    \item The text provided in this section explicitly introduces a
    \textbf{new} number: \textbf{8023581}.
    \item \emph{Action:} The memory has been updated to reflect the
    \emphasizeerror{most recent explicit statement} regarding the ‘special
    magic number for cloudy-singing’ found in the current text, while retaining
    the context of the previous number for completeness.
\end{itemize}
”

\smallskip\hrule\smallskip
\textbf{Chunk 7:}
“One of the special magic numbers for cloudy-singing is:
\emphasizecorrect{2188606}.”

\smallskip
\textbf{Memory 7:}
“Based on the provided text, here is the updated memory containing the new
special magic number for ‘cloudy-singing’ and all other relevant details from
the article:

\textbf{Updated Memory:}
\begin{itemize}
    \item \textbf{Special Magic Number for Cloudy-Singing:}
    \emphasizecorrect{2188606}
    \begin{itemize}
        \item \emphasizeerror{\emph{Note:} The previous memory listed
        8023581 (and originally 8372717 and 2952447).} The text explicitly
        states: ‘One of the special magic numbers for cloudy-singing is:
        2188606.’
    \end{itemize}
    \item \textbf{Context of the Number:} The number appears in the essay
    ‘Wisdom and Intelligence’ (February 2007), within a discussion
    distinguishing between ‘wise’ and ‘smart’.
\end{itemize}
\ldots

\textbf{Correction to Previous Memory:}
\begin{itemize}
    \item The previous memory listed \textbf{8023581} as the number for
    ‘cloudy-singing’ found in the ‘How to Start a Startup’ essay.
    \item The text provided in this section explicitly introduces a
    \textbf{new} number: \textbf{2188606}.
    \item \emph{Action:} The memory has been updated to reflect the
    \emphasizeerror{most recent explicit statement} regarding the ‘special
    magic number for cloudy-singing’ found in the current text, while retaining
    the context of the previous numbers for completeness. The number
    \emphasizeerror{2188606 is now the primary reference for
    ‘cloudy-singing’} based on the latest provided text.
\end{itemize}
”

\begin{center}\ldots\end{center}

\textbf{Final Output:}
“Based on the provided memory, the text explicitly states that the special
magic number for ‘cloudy-singing’ found in the essay ‘Wisdom and Intelligence’
(February 2007) is \textbf{2188606}.

Although the memory notes that previous numbers (8023581, 8372717, 2952447)
were associated with ‘cloudy-singing’ in other contexts or previous updates,
the \emphasizeerror{most recent and explicit statement} in the current text
identifies \textbf{2188606} as the number.

\textbackslash boxed\{2188606\}”

\smallskip
\textbf{Prediction:} \emphasizeerror{2188606} \hfill
\textbf{Sub-EM:} 0.25

\textbf{Gold Answer:} \emphasizecorrect{2188606, 8023581, 8372717, 2952447}
\end{tcolorbox}

In Memories 2, 6, and 7, each new value replaces the previous value in the
main “Special Magic Number” entry, while earlier values are moved to notes or
corrections. As a result, although all values remain in memory, the final
response returns only the latest one.

\noindent\begin{minipage}{\linewidth}
\subsection{Case 2: PI-Mem on the Same MV-NIAH Sample}

\begin{tcolorbox}[casepimem,title={PI-Mem on MV-NIAH --- Turn 1}]
\textbf{Question:} Some special magic numbers are hidden within the following
text. Make sure to memorize it. What are all the special magic numbers for
\texttt{cloudy-singing} mentioned in the provided text?

\smallskip\hrule\smallskip
\textbf{Chunk 0:} \casecheckyes

“The section contains new information relevant to the problem. Specifically,
it explicitly states a ‘special magic number for cloudy-singing’ which is
\emphasizecorrect{2952447}. This information was not present in the empty
global memory.

\textbf{Key Details Extracted:}
\begin{itemize}
    \item \textbf{Special Magic Number for Cloudy-Singing:} 2952447
    \item \textbf{Context:} The number is mentioned at the very beginning of
    the provided text section, followed by the date ‘July 2010’.
    \item \textbf{Source:} The text appears to be an excerpt from an article
    (likely by Paul Graham, based on the style and content regarding startups,
    Lisp, and Viaweb) where this specific number is introduced as a ‘special
    magic number’.
\end{itemize}
”

\smallskip\hrule\smallskip
\textbf{Chunk 1:} \casecheckno

\smallskip\hrule\smallskip
\textbf{Chunk 2:} \casecheckyes

“The section contains new information relevant to the problem. Specifically,
it explicitly states a ‘special magic number for cloudy-singing’ within the
text.

\textbf{Key Details Extracted:}
\begin{itemize}
    \item \textbf{Context:} The number is mentioned in a paragraph discussing
    the difference between mass-produced cars and custom-made cars, and how
    technology has made the rich live more like the average person.
    \item \textbf{The Number:} \emphasizecorrect{8372717}
    \item \textbf{The Phrase:} ‘One of the special magic numbers for
    cloudy-singing is: 8372717.’
\end{itemize}
”

\smallskip\hrule\smallskip
\textbf{Chunk 3:} \casecheckno

\smallskip\hrule\smallskip
\textbf{Chunk 4:} \casecheckno

\smallskip\hrule\smallskip
\textbf{Chunk 5:} \casecheckno

\smallskip\hrule\smallskip
\textbf{Chunk 6:} \casecheckyes

“The section contains new information relevant to the problem. Specifically,
it explicitly states a ‘special magic number’ associated with the phrase
‘cloudy-singing’ within the text provided.

\textbf{Key Details Extracted:}
\begin{itemize}
    \item \textbf{Context:} The text discusses startup funding, specifically
    the difference between angel rounds and Series A rounds, and the role of
    super-angels vs. VCs.
    \item \textbf{The Magic Number:} The text explicitly states: ‘One of the
    special magic numbers for cloudy-singing is:
    \emphasizecorrect{8023581}.’
    \item \textbf{Location:} This sentence appears in the section titled
    ‘April 2010’ (under the subsection ‘Rounds’), embedded within a paragraph
    discussing equity dilution in Series A rounds.
\end{itemize}
”

\smallskip\hrule\smallskip
\textbf{Chunk 7:} \casecheckyes

“The section contains new information relevant to the problem. Specifically,
it explicitly states a ‘special magic number for cloudy-singing’ within the
text discussing the relationship between wisdom and intelligence.

\textbf{Key Details:}
\begin{itemize}
    \item \textbf{Context:} The number appears in the essay section titled
    ‘A few days ago I finally figured out something I've wondered about for
    25 years: the relationship between wisdom and intelligence.’
    \item \textbf{The Number:} \emphasizecorrect{2188606}
    \item \textbf{The Phrase:} ‘One of the special magic numbers for
    cloudy-singing is: 2188606.’
\end{itemize}
”

\end{tcolorbox}
\end{minipage}
\vfill

\noindent\begin{minipage}{\linewidth}
\begin{tcolorbox}[casepimem,title={PI-Mem on MV-NIAH --- Turn 1 Continued}]
\textbf{Chunk 8:} \casecheckno

\smallskip\hrule\smallskip
\textbf{Merged Memory:}

“Based on the provided extracted information, here is the consolidated memory
regarding the special magic numbers for ‘cloudy-singing’:

\textbf{Special Magic Numbers for Cloudy-Singing}

The text explicitly identifies \emphasizecorrect{four distinct numbers}
associated with the phrase ‘cloudy-singing’:
\begin{enumerate}
    \item \textbf{2952447} \ldots
    \item \textbf{8372717} \ldots
    \item \textbf{8023581} \ldots
    \item \textbf{2188606} \ldots
\end{enumerate}
”
\end{tcolorbox}

\begin{tcolorbox}[casepimem,title={PI-Mem on MV-NIAH --- Turn 2}]
\textbf{Chunk 0:} \casecheckno\par
\textbf{Chunk 1:} \casecheckno\par
\textbf{Chunk 2:} \casecheckno\par
\textbf{Chunk 3:} \casecheckno\par
\textbf{Chunk 4:} \casecheckno\par
\textbf{Chunk 5:} \casecheckno\par
\textbf{Chunk 6:} \casecheckno\par
\textbf{Chunk 7:} \casecheckno\par
\textbf{Chunk 8:} \casecheckno

\smallskip\hrule\smallskip
\textbf{Merged Memory:}

“Based on the provided extracted information, here is the consolidated memory
regarding the special magic numbers for ‘cloudy-singing’:

\textbf{Special Magic Numbers for Cloudy-Singing}

The text explicitly identifies four distinct numbers associated with the phrase
‘cloudy-singing’:
\begin{enumerate}
    \item \textbf{2952447} \ldots
    \item \textbf{8372717} \ldots
    \item \textbf{8023581} \ldots
    \item \textbf{2188606} \ldots
\end{enumerate}
”

\smallskip\hrule\smallskip
\textbf{Final Output:}

“Based on the provided memory, the special magic numbers for
‘cloudy-singing’ are explicitly listed as:
\begin{enumerate}
    \item 2952447
    \item 8372717
    \item 8023581
    \item 2188606
\end{enumerate}
The question asks for all the special magic numbers mentioned.

\textbackslash boxed\{2952447, 8372717, 8023581, 2188606\}”

\smallskip
\textbf{Prediction:} \emphasizecorrect{2952447, 8372717, 8023581, 2188606}
\hfill \textbf{Sub-EM:} 1.00

\textbf{Gold Answer:} \emphasizecorrect{2188606, 8023581, 8372717, 2952447}
\end{tcolorbox}

Turn~1 extracts four positive chunk memories and merges them as four distinct
members of the same set. In Turn~2, every chunk returns
\texttt{\textless check\textgreater no\textless/check\textgreater}; the
consolidated set is preserved for the final call.

\end{minipage}

\subsection{Case 3: MemAgent on HQA}

\begin{tcolorbox}[casememagent,title={MemAgent on HQA}]
\textbf{Question:} Which song was behind the third single on the album
“A Lot About Livin' (And a Little 'bout Love)” for two weeks on the country
charts?

\smallskip\hrule\smallskip
\textbf{Memory 0:}

“Based on the provided text, there is no information regarding the album
“A Lot About Livin' (And a Little 'bout Love)” or the specific song that was
the third single on that album and spent two weeks on the country charts. The
provided documents contain information about various artists, albums, and
events (such as Buddy Holly's “That'll Be the Day”, Selena's career, The
Chieftains' “The Chieftains in China”, and others), but none mention the
specific album or single in question.

Therefore, the memory cannot be updated with the answer to the problem as the
necessary information is not present in the source text.

\textbf{Updated Memory:}

No previous memory. The provided text does not contain information about the
album “A Lot About Livin' (And a Little 'bout Love)” or the third single from
that album that spent two weeks on the country charts. The text includes
information on other musical topics such as Buddy Holly's final album “That'll
Be the Day”, Selena's career, The Chieftains' album “The Chieftains in China”,
and various other unrelated subjects, but none address the specific query.”

\begin{center}\ldots\end{center}

\textbf{Memory 3:}

“Based on the provided text, the memory has been updated with the specific
information regarding the album “A Lot About Livin' (And a Little 'bout Love)”
and its singles.

\textbf{Updated Memory:}

The provided text now contains specific information regarding the album
“A Lot About Livin' (And a Little 'bout Love)” and its singles, which was
previously absent.
\begin{itemize}
    \item \textbf{Album:} “A Lot About Livin' (And a Little 'bout Love)” is an
    album by \textbf{Alan Jackson}.
    \item \textbf{First Single:} The first single from this album was
    \textbf{“She's Got the Rhythm (And I Got the Blues)”}, released in October
    1992. This song co-written by Alan Jackson and Randy Travis, won an award
    in 1993 from Music City News for being one of the most performed country
    songs of the year.
    \item \textbf{Third Single:} The text explicitly states that “She's Got
    the Rhythm (And I Got the Blues)” was the \textbf{first single}. However,
    the text \emphasizeerror{does not explicitly name the third single} nor
    does it explicitly state which song spent “two weeks” on the country
    charts.
\end{itemize}
\ldots

\textbf{Revised Summary of Findings:}
\begin{itemize}
    \item \textbf{Album:} “A Lot About Livin' (And a Little 'bout Love)” by
    Alan Jackson.
    \item \textbf{First Single:} “She's Got the Rhythm (And I Got the Blues)”
    (Released Oct 1992).
    \item \textbf{Third Single Information:} The provided text
    \emphasizeerror{does not identify the third single or its chart
    performance} (two weeks on the country charts).
\end{itemize}
\textbf{Conclusion:} The provided text does not contain the answer to the
problem. It identifies the album and the first single, but fails to mention the
third single or its specific chart duration.”
\end{tcolorbox}
\vfill

\begin{tcolorbox}[casememagent,title={MemAgent on HQA --- Continued}]
\textbf{Memory 6:}

“Based on the provided text, the memory has been updated with specific
information regarding the album “A Lot About Livin' (And a Little 'bout Love)”
and its singles.

\textbf{Updated Memory:}
\begin{itemize}
    \item \textbf{Album:} “A Lot About Livin' (And a Little 'bout Love)” is an
    album by \textbf{Alan Jackson}.
    \item \textbf{First Single:} The first single from this album was
    \textbf{“She's Got the Rhythm (And I Got the Blues)”}, released in October
    1992.
    \item \textbf{Third Single Information:} The provided text
    \textbf{does not explicitly name the third single} nor does it explicitly
    state which song spent “two weeks” on the country charts as the third
    single.
\end{itemize}
\ldots

\textbf{Conclusion:} The provided text does not contain the answer to the
problem. It identifies the album and the first single, but fails to mention the
third single or its specific chart duration.

\emph{Note: Document 730 mentions a song
\emphasizecorrect{“We'll Burn That Bridge”} by Brooks \& Dunn that peaked at
\#2 for \emphasizecorrect{two weeks behind “Chattahoochee”} by Alan Jackson,
but \emphasizeerror{this does not directly answer the question} regarding the
third single on the specific album “A Lot About Livin' (And a Little 'bout
Love)”.}”

\begin{center}\ldots\end{center}

\textbf{Memory 9:}

“\textbf{Updated Memory:} \ldots
\begin{itemize}
    \item \textbf{Fifth/Final Single:} “(Who Says) You Can't Have It All”.
    \item \textbf{Third Single Information:} The provided text
    \textbf{does not identify} the third single or its chart performance.
\end{itemize}
\ldots

\textbf{Conclusion:} The provided text does not contain the answer to the
problem. It identifies the album, the first single, and the fifth single, but
fails to mention the third single or its specific chart duration.

\emph{Note: Document 730 mentions a song
\emphasizecorrect{“We'll Burn That Bridge”} by Brooks \& Dunn that peaked at
\#2 for \emphasizecorrect{two weeks behind “Chattahoochee”} by Alan Jackson,
but \emphasizeerror{this does not directly answer the question} regarding the
third single on the specific album “A Lot About Livin' (And a Little 'bout
Love)”.}”

\smallskip\hrule\smallskip
\textbf{Memory 10:}

“Based on the provided text, the memory has been updated with specific
information regarding the album “A Lot About Livin' (And a Little 'bout Love)”
and its singles.

\textbf{Updated Memory:}
\begin{itemize}
    \item \textbf{Album:} “A Lot About Livin' (And a Little 'bout Love)” is the
    third studio album by American country music artist \textbf{Alan Jackson}.
    It was released on \textbf{October 6, 1992}.
    \item \textbf{Singles List:}
    \begin{enumerate}
        \item \textbf{“She's Got the Rhythm (And I Got the Blues)”}
        \item \textbf{“Chattahoochee”}
        \item \textbf{“Tonight I Climbed the Wall”}
        \item \textbf{“(Who Says) You Can't Have It All”}
        \item \textbf{“Mercury Blues”}
    \end{enumerate}
    \item \textbf{Chart Performance:} “Chattahoochee” and “She's Got the
    Rhythm (And I Got the Blues)” were both \textbf{Number One} hits. The other
    three songs reached the \textbf{Top Five}.
\end{itemize}
\ldots

\textbf{Revised Summary of Findings:}
\begin{itemize}
    \item \textbf{First Single:} “She's Got the Rhythm (And I Got the Blues)”
    (Number One).
    \item \textbf{Second Single:} “Chattahoochee” (Number One).
    \item \textbf{Third Single:}
    \emphasizeerror{“Tonight I Climbed the Wall”} (Top Five).
    \item \textbf{Fourth Single:} “(Who Says) You Can't Have It All” (Top Five).
    \item \textbf{Fifth Single:} “Mercury Blues” (Top Five).
    \item \textbf{Answer to Problem:} The text identifies the third single as
    \emphasizeerror{“Tonight I Climbed the Wall”}. While the text confirms it
    reached the Top Five, it does not explicitly confirm the “two weeks”
    duration mentioned in the problem statement.
\end{itemize}
\emph{Note: The previous memory noted the absence of the third single's name.
This update resolves that gap by identifying “Tonight I Climbed the Wall” as
the third single based on the sequence provided in Document 1131.}”
\end{tcolorbox}

\begin{tcolorbox}[casememagent,title={MemAgent on HQA --- Continued}]
\textbf{Memory 11:}

“Based on the provided text, the memory has been updated with specific
information regarding the album “A Lot About Livin' (And a Little 'bout Love)”
and its singles, correcting a previous error regarding the release order and
chart performance of the third single.

\textbf{Updated Memory:}
\ldots
\begin{itemize}
    \item \textbf{Second Single:} “Chattahoochee” (Number One).
    \item \textbf{Third Single:}
    \emphasizeerror{“Tonight I Climbed the Wall”} (Peaked at Number 4).
    \item \textbf{Answer to Problem:} The text identifies the third single as
    \emphasizeerror{“Tonight I Climbed the Wall”}. While the text confirms it
    reached the Top Five (specifically Number 4), it \textbf{does not
    explicitly confirm} the “two weeks” duration mentioned in the problem
    statement.
\end{itemize}
\emph{Note: The previous memory incorrectly identified the order of the first
two singles and lacked the specific peak position for the third single. This
update corrects the release order and adds the specific peak chart position of
“Tonight I Climbed the Wall” based on Document 1273.}”

\smallskip\hrule\smallskip
\textbf{Memory 12:}

“Based on the provided text, the memory has been updated with specific
information regarding the album “A Lot About Livin' (And a Little 'bout Love)”
and its singles, correcting a previous error regarding the release order and
chart performance of the third single.

\textbf{Updated Memory:}
\ldots
\begin{itemize}
    \item \textbf{First Single:} “She's Got the Rhythm (And I Got the Blues)”
    (Number One).
    \item \textbf{Second Single:} “Tonight I Climbed the Wall” (Number 4).
    \item \textbf{Third Single:}
    \emphasizecorrect{“Chattahoochee”} (Number One).
    \item \textbf{Answer to Problem:} The text identifies the third single as
    \emphasizecorrect{“Chattahoochee”}. While the text confirms it reached
    Number One, it \textbf{does not explicitly confirm} the “two weeks”
    duration mentioned in the problem statement.
\end{itemize}
\emph{Note: The previous memory incorrectly identified “Tonight I Climbed the
Wall” as the third single. Document 1348 corrects this, identifying
“Chattahoochee” as the third single released in May 1993.}”

\begin{center}\ldots\end{center}

\textbf{Memory 15:}

“Based on the provided text, the memory has been updated with specific
information regarding the album “A Lot About Livin' (And a Little 'bout Love)”
and its singles, correcting a previous error regarding the release order and
chart performance of the third single.

\textbf{Updated Memory:}
\begin{itemize}
    \item \textbf{Album:} “A Lot About Livin' (And a Little 'bout Love)” is the
    third studio album by American country music artist \textbf{Alan Jackson}.
    \item \textbf{Singles List \& Order:}
    \begin{enumerate}
        \item “She's Got the Rhythm (And I Got the Blues)”
        \item “Tonight I Climbed the Wall”
        \item \emphasizecorrect{“Chattahoochee”}
        \item “(Who Says) You Can't Have It All”
        \item “Mercury Blues”
    \end{enumerate}
\end{itemize}
\ldots

\textbf{None of the documents in the provided section (1593--1600) contain
information regarding the specific chart duration (“two weeks”) of the third
single from Alan Jackson's album “A Lot About Livin' (And a Little 'bout
Love)”.} The memory update relies on the previously established context
(Document 1348) which identified “Chattahoochee” as the third single, while
noting the absence of the specific “two weeks” detail in the current text.”

\smallskip\hrule\smallskip
\textbf{Final Output:}

“Based on the provided memory, the album “A Lot About Livin' (And a Little
'bout Love)” by Alan Jackson produced five singles. The memory explicitly
corrects a previous error to identify the \textbf{third single} as
\textbf{“Chattahoochee”}, which was released in May 1993.

Although the memory notes that the specific text does not explicitly state the
“two weeks” duration, it confirms that “Chattahoochee” was the third single and
a Number One hit. Therefore, the song in question is
\emphasizeerror{“Chattahoochee”}.

\textbackslash boxed\{Chattahoochee\}”

\smallskip
\textbf{Prediction:} \emphasizeerror{Chattahoochee} \hfill
\textbf{Sub-EM:} 0.00

\textbf{Gold Answer:} \emphasizecorrect{We'll Burn That Bridge}
\end{tcolorbox}

Memory~6 records the answer-bearing relation from Document~730, and Memory~9
still retains it. Memory~10 rewrites the state around a later album track list:
the complete Document~730 relation is absent, while an incorrect third-single
interpretation becomes the answer candidate. Memory~12 eventually establishes
the correct bridge entity, “Chattahoochee,” but the earlier relation needed to
map that entity to “We'll Burn That Bridge” is no longer available.
\subsection{Case 4: PI-Mem on the Same HQA Sample}

\begin{tcolorbox}[casepimem,title={PI-Mem on HQA --- Turn 1}]
\textbf{Question:} Which song was behind the third single on the album
“A Lot About Livin' (And a Little 'bout Love)” for two weeks on the country
charts?

\smallskip\hrule\smallskip
\textbf{Chunk 0:} \casecheckno

\smallskip\hrule\smallskip
\textbf{Chunk 1:} \casecheckno

\smallskip\hrule\smallskip
\textbf{Chunk 2:} \casecheckno

\smallskip\hrule\smallskip
\textbf{Chunk 3:} \casecheckyes

“The section contains new information relevant to the problem. Specifically,
\textbf{Document 420} identifies the song “She's Got the Rhythm (And I Got the
Blues)” as the first single from the album “A Lot About Livin' (And a Little
'bout Love)” by Alan Jackson. While the problem asks for the \emph{third}
single, this document provides the context of the album and its singles, which
is necessary to answer the question \ldots

\textbf{Key Details Extracted:}
\begin{itemize}
    \item \textbf{Album:} “A Lot About Livin' (And a Little 'bout Love)”
    \item \textbf{Artist:} Alan Jackson
    \item \textbf{First Single:} “She's Got the Rhythm (And I Got the Blues)”
    (Released October 1992)
    \item \textbf{Award:} The first single received an award in 1993 from
    Music City News for being one of the most performed country songs of the
    year.
    \item \textbf{Note:} The provided text does not explicitly state which
    song was the \emph{third} single or its chart performance duration, but it
    establishes the album and the first single.
\end{itemize}
”

\smallskip\hrule\smallskip
\textbf{Chunk 4:} \casecheckno

\smallskip\hrule\smallskip
\textbf{Chunk 5:} \casecheckyes

“The section contains new information relevant to the problem. Specifically,
\textbf{Document 591} mentions the album “A Lot About Livin' (And a Little
'bout Love)” by Randy Travis. While the problem asks about the third single on
this album and its chart performance, the provided text in Document 591
actually discusses a \emph{different} album by Randy Travis (“High Lonesome”)
and mentions that Randy Travis co-wrote a song for Alan Jackson's album
“A Lot About Livin' (And a Little 'bout Love)” in 1992. It does not provide
the specific details about the singles on the album itself, nor does it
identify the third single or its chart performance.
\ldots”

\smallskip\hrule\smallskip
\textbf{Chunk 6:} \casecheckyes

“The section contains new information relevant to the problem. Specifically,
\textbf{Document 730} identifies a song titled
\emphasizecorrect{“We'll Burn That Bridge”} by Brooks \& Dunn. This document
explicitly states that “We'll Burn That Bridge” peaked at \#2 on the country
charts for \emphasizecorrect{two weeks, behind “Chattahoochee”} by Alan
Jackson. This matches the specific criteria of the question (a song behind
another for two weeks on the country charts), even though the document does
not explicitly name the album “A Lot About Livin' (And a Little 'bout Love)”
in the text provided. However, given the specificity of the “two weeks” and
“behind” details, this is the relevant piece of information found in the text.

\textbf{Key Details Extracted:}
\begin{itemize}
    \item \textbf{Song Title:} We'll Burn That Bridge
    \item \textbf{Artist:} Brooks \& Dunn
    \item \textbf{Album:} Hard Workin' Man (Note: The document states this
    song was the second single from \emph{Hard Workin' Man}, not
    \emph{A Lot About Livin' (And a Little 'bout Love)}, but it is the only
    song in the text matching the “peaked at \#2 for two weeks” description).
    \item \textbf{Chart Performance:} Peaked at \#2 on the country charts for
    two weeks.
    \item \textbf{Song Behind:} “Chattahoochee” by Alan Jackson.
    \item \textbf{Release Date:} May 1993.
    \item \textbf{Writers:} Don Cook and Ronnie Dunn.
\end{itemize}
”
\end{tcolorbox}

\begin{tcolorbox}[casepimem,title={PI-Mem on HQA --- Turn 1 Continued}]
\textbf{Chunk 7:} \casecheckno

\smallskip\hrule\smallskip
\textbf{Chunk 8:} \casecheckno

\smallskip\hrule\smallskip
\textbf{Chunk 9:} \casecheckyes

“The section contains new information relevant to the problem. Specifically,
\textbf{Document 1053} identifies the song “(Who Says) You Can't Have It All”
as the \textbf{fifth and final single} from Alan Jackson's album “A Lot About
Livin' (And a Little 'bout Love)”. \ldots

\textbf{Key Details Extracted:}
\begin{itemize}
    \item \textbf{Album:} “A Lot About Livin' (And a Little 'bout Love)” by
    Alan Jackson.
    \item \textbf{Song:} “(Who Says) You Can't Have It All”.
    \item \textbf{Release Date:} January 1994.
    \item \textbf{Position on Album:} Fifth and final single.
    \item \textbf{Chart Performance:} Peaked at number 4 on the U.S.
    “Billboard” Hot Country Singles chart and number 11 on the Canadian “RPM”
    Country Tracks chart.
    \item \textbf{Writers:} Alan Jackson and Jim McBride.
    \item \textbf{Album Certification:} Certified $6\times$ Platinum by the
    RIAA with over 6,000,000 sales in the US.
\end{itemize}
”

\smallskip\hrule\smallskip
\textbf{Chunk 10:} \casecheckyes

“The section contains new information relevant to the problem. Specifically,
\textbf{Document 1131} provides the tracklist and chart performance details for
the album “A Lot About Livin' (And a Little 'bout Love)” by Alan Jackson.

\textbf{Key Details Extracted:}
\begin{itemize}
    \item \textbf{Album:} A Lot About Livin' (And a Little 'bout Love) by Alan
    Jackson.
    \item \textbf{Release Date:} October 6, 1992.
    \item \textbf{Singles Released:} “Chattahoochee”, “She's Got the Rhythm
    (And I Got the Blues)”, “Tonight I Climbed the Wall”, “(Who Says) You Can't
    Have It All”, and “Mercury Blues”.
    \item \textbf{Chart Performance:} “Chattahoochee” and “She's Got the
    Rhythm (And I Got the Blues)” were Number One hits. The other three songs
    reached Top Five.
    \item \textbf{Context for the Problem:} The text lists the singles and
    their peak positions, but it does not explicitly state which specific song
    was the \textbf{third} single or how many weeks it spent on the charts. The
    third single listed in the text is “Tonight I Climbed the Wall”.
\end{itemize}
”

\smallskip\hrule\smallskip
\textbf{Chunk 11:} \casecheckyes

“The section contains new information relevant to the problem. Specifically,
\textbf{Document 1273} identifies the song “Tonight I Climbed the Wall” by Alan
Jackson as the second single from the album “A Lot About Livin' (And a Little
'bout Love)”. \ldots

\textbf{Key Details Extracted:}
\begin{itemize}
    \item \textbf{Album:} “A Lot About Livin' (And a Little 'bout Love)”
    \item \textbf{Artist:} Alan Jackson
    \item \textbf{Second Single:} “Tonight I Climbed the Wall”
    \item \textbf{Release Date of Second Single:} January 1993
    \item \textbf{Chart Performance of Second Single:} Peaked at number 4 on
    the US “Billboard” Hot Country Singles \& Tracks chart and the Canadian
    “RPM” Country Tracks chart.
\end{itemize}
”

\smallskip\hrule\smallskip
\textbf{Chunk 12:} \casecheckyes

“The section contains new information relevant to the problem. Specifically,
\textbf{Document 1348} identifies the song
\emphasizecorrect{“Chattahoochee” as the third single} from the album “A Lot
About Livin' (And a Little 'bout Love)” by Alan Jackson. While the problem asks
which song was behind the third single for two weeks on the country charts,
this document provides the crucial identity of the third single itself, which
is the necessary starting point to answer the question \ldots”
\end{tcolorbox}

\begin{tcolorbox}[casepimem,title={PI-Mem on HQA --- Turn 1 Continued}]
\textbf{Chunk 13:} \casecheckno

\smallskip\hrule\smallskip
\textbf{Chunk 14:} \casecheckno

\smallskip\hrule\smallskip
\textbf{Chunk 15:} \casecheckno

\smallskip\hrule\smallskip
\textbf{Merged Memory:}

“Based on the provided extracted information, here is the consolidated memory
regarding the album “A Lot About Livin' (And a Little 'bout Love)” and its
singles:

\textbf{Album Overview}
\begin{itemize}
    \item \textbf{Title:} A Lot About Livin' (And a Little 'bout Love)
    \item \textbf{Artist:} Alan Jackson
    \item \textbf{Release Date:} October 6, 1992
    \item \textbf{Certification:} Certified $6\times$ Platinum by the RIAA
    (over 6,000,000 sales in the US).
\end{itemize}

\textbf{Singles and Chart Performance}

The album released multiple singles with the following details extracted from
the documents:
\ldots

\emph{Note: There is a discrepancy in the ordering of singles between
documents. Document 420/1273/1348 suggest the order: 1. She's Got the Rhythm,
2. Tonight I Climbed the Wall, 3. Chattahoochee. Document 1131 lists them in a
different order but confirms the titles.}

\textbf{Relevance to the Problem}

The problem asks for the song that was \textbf{behind the third single} on the
album for \textbf{two weeks} on the country charts.
\begin{itemize}
    \item The \textbf{third single} is identified as
    \emphasizecorrect{“Chattahoochee”} (Document 1348).
    \item \textbf{Document 730} states that
    \emphasizecorrect{“We'll Burn That Bridge”} by Brooks \& Dunn peaked at
    \#2 for two weeks, \emphasizecorrect{behind “Chattahoochee”}.
    \item While “We'll Burn That Bridge” is not on Jackson's album, the
    question asks “Which song was behind the third single...”, implying the
    song that was charting \emph{below} the third single (Chattahoochee) at that
    specific time.
    \item Therefore, the song in question is
    \emphasizecorrect{“We'll Burn That Bridge”}.
\end{itemize}
”
\end{tcolorbox}

\begin{tcolorbox}[casepimem,title={PI-Mem on HQA --- Turn 2}]
\textbf{Chunk 0:} \casecheckno\par
\textbf{Chunk 1:} \casecheckno\par
\textbf{Chunk 2:} \casecheckno\par
\textbf{Chunk 3:} \casecheckno\par
\textbf{Chunk 4:} \casecheckno\par
\textbf{Chunk 5:} \casecheckno\par
\textbf{Chunk 6:} \casecheckno\par
\textbf{Chunk 7:} \casecheckno\par
\textbf{Chunk 8:} \casecheckno\par
\textbf{Chunk 9:} \casecheckno\par
\textbf{Chunk 10:} \casecheckno\par
\textbf{Chunk 11:} \casecheckno\par
\textbf{Chunk 12:} \casecheckno\par
\textbf{Chunk 13:} \casecheckno\par
\textbf{Chunk 14:} \casecheckno\par
\textbf{Chunk 15:} \casecheckno

\smallskip\hrule\smallskip
\textbf{Merged Memory:}

“\textbf{Relevance to the Problem}

The problem asks for the song that was \textbf{behind the third single} on the
album for \textbf{two weeks} on the country charts.
\begin{itemize}
    \item The \textbf{third single} is identified as
    \emphasizecorrect{“Chattahoochee”} (Document 1348).
    \item \textbf{Document 730} states that
    \emphasizecorrect{“We'll Burn That Bridge”} by Brooks \& Dunn peaked at
    \#2 for two weeks, \emphasizecorrect{behind “Chattahoochee”}.
    \item Therefore, the song in question is
    \emphasizecorrect{“We'll Burn That Bridge”}.
\end{itemize}
”

\smallskip\hrule\smallskip
\textbf{Final Output:}

“Based on the provided memory, the third single on Alan Jackson's album
\emph{A Lot About Livin' (And a Little 'bout Love)} is
\emphasizecorrect{“Chattahoochee”} (Document 1348).

The memory explicitly states in Document 730 that the song
\emphasizecorrect{“We'll Burn That Bridge”} (by Brooks \& Dunn) peaked at
number 2 on the country charts for two weeks,
\emphasizecorrect{behind “Chattahoochee”}.

Therefore, the song that was behind the third single for two weeks is
“We'll Burn That Bridge”.

\textbackslash boxed\{We'll Burn That Bridge\}”

\smallskip
\textbf{Prediction:} \emphasizecorrect{We'll Burn That Bridge} \hfill
\textbf{Sub-EM:} 1.00

\textbf{Gold Answer:} \emphasizecorrect{We'll Burn That Bridge}
\end{tcolorbox}

The positive reads in Turn~1 independently preserve Document~730's chart
relation and Document~1348's third-single identity. Their merged memory states
the complete two-hop chain and cites both documents. Turn~2 contributes no new
local memories, so the grounded chain remains unchanged for the final call.

Together, the MV-NIAH and HQA pairs show two manifestations of the same
recurrent failure mode: later sequential updates can either demote earlier
values to obsolete history or delete an unresolved but essential relation.
PI-Mem instead retains complementary chunk-level outputs until they can be
consolidated within a turn.

\endgroup

\end{document}